\documentclass[sigconf]{acmart}
\usepackage{amsmath}
\usepackage{booktabs}
\usepackage{graphicx}
\usepackage{multirow}
\usepackage{array}
\usepackage{adjustbox}
\usepackage{enumitem}
\usepackage{booktabs,multirow,adjustbox,colortbl,xcolor}
\AtBeginDocument{%
 }

\setcopyright{none}
\renewcommand\footnotetextcopyrightpermission[1]{}

\acmConference[KDD '27]{The 33rd ACM SIGKDD Conference on Knowledge Discovery and Data Mining}{August 1--5, 2027}{San Jose, CA, USA}

\begin{document}

\title{AutoSND: From Execution Evidence to Structural Policies for Automated Network Dismantling Heuristic Discovery}

\author{Zhijing Hu}
\orcid{0009-0003-5413-7483}
\affiliation{%
 \institution{College of Systems Engineering, National University of Defense Technology}
 \city{Changsha}
 \country{China}}
 \email{huzhijing@nudt.edu.cn}

\author{Changjun Fan}
\correspondingauthor
\affiliation{%
 \institution{College of Systems Engineering, National University of Defense Technology}
 \city{Changsha}
 \country{China}}
 \email{fanchangjun09@163.com}

\author{Yufan Deng}
\affiliation{%
 \institution{College of Systems Engineering, National University of Defense Technology}
 \city{Changsha}
 \country{China}}
 \email{dengyufan24@nudt.edu.cn}

\author{Zhiguang Cao}
\affiliation{%
 \institution{School of Computing and Information Systems, Singapore Management University}
 % \city{Singapore}
 \country{Singapore}}
 \email{zgcao@smu.edu.sg}

\renewcommand{\shortauthors}{Hu et al.}

\begin{abstract}
Network dismantling is fundamental to analyzing the robustness and vulnerability of complex systems, yet practical heuristics must balance effectiveness and computational efficiency, and are usually designed manually by researchers. Existing large language model based automatic heuristic design methods can generate and screen candidates, yet they have difficulty further transforming candidate quality or failure states during execution into structural-level guidance for subsequent generation. We propose AutoSND, a three stage tree search framework for complete network dismantling programs. Stage I broadly explores from simple heuristics and archives execution evidence. Stage II compiles candidate records into structural policies concerning local signals, neighborhood access, and state update ranges. Stage III continues tree search conditioned on these policies and obtains the final quality prioritized and speed prioritized candidates, AutoSND-Q/S. Experiments on 12 real world networks and 3 large real world networks show that AutoSND achieves better search performance and stability and discovers more competitive and structurally interpretable network dismantling programs. The final candidates form an interpretable structure that uses residual degree as the backbone, adjusts node order with bounded local signals, and restricts the state update range. Code is available at \url{https://github.com/MirrorNew/AutoSND}.
\end{abstract}

\makeatletter
\gdef\@concepts{\textbullet\ \textbf{Computing methodologies} \textrightarrow\ \textbf{Heuristic function construction}. \textbullet\ \textbf{Networks} \textrightarrow\ \textbf{Network algorithms}. \textbullet\ \textbf{Theory of computation} \textrightarrow\ \textbf{Graph algorithms analysis}.}
\makeatother


\maketitle

\section{Introduction}

Network dismantling identifies and removes critical nodes to rapidly weaken connectivity, making it fundamental to the robustness analysis of complex networks~\cite{morone2015,braunstein2016,zdeborova2016,ren2019}. The key scientific question is which locally observable structural mechanisms govern global fragmentation across networks with different formation processes. Existing heuristics generally need to balance dismantling effectiveness and computational efficiency, and they use local structures such as node degree, core structure, neighborhood connections, or message passing to generate node removal sequences at low cost~\cite{morone2015,braunstein2016,zdeborova2016,ren2019,fan2020,grassia2021,zhang2022}. However, their structural signals, access patterns, and update mechanisms still mainly rely on expert knowledge and manual design.

Large language model based automatic heuristic design (LLM based AHD) offers an alternative. FunSearch, EoH, ReEvo, and MCTS-AHD combine LLM code generation with evolutionary search, reflection, or tree search~\cite{romeraparedes2024,liu2024eoh,ye2024,zheng2025}, while ReEvo and HiFo-Prompt summarize historical experience to guide later generation~\cite{ye2024,mao2024}. In network science, LLM4CN evolves static node scoring functions from a manually designed initial population through LLM based crossover and mutation~\cite{zhang2025llm4cn}, demonstrating the potential of LLM based AHD for network dismantling.

\begin{figure*}[t]
 \centering
 \includegraphics[width=0.8\textwidth]{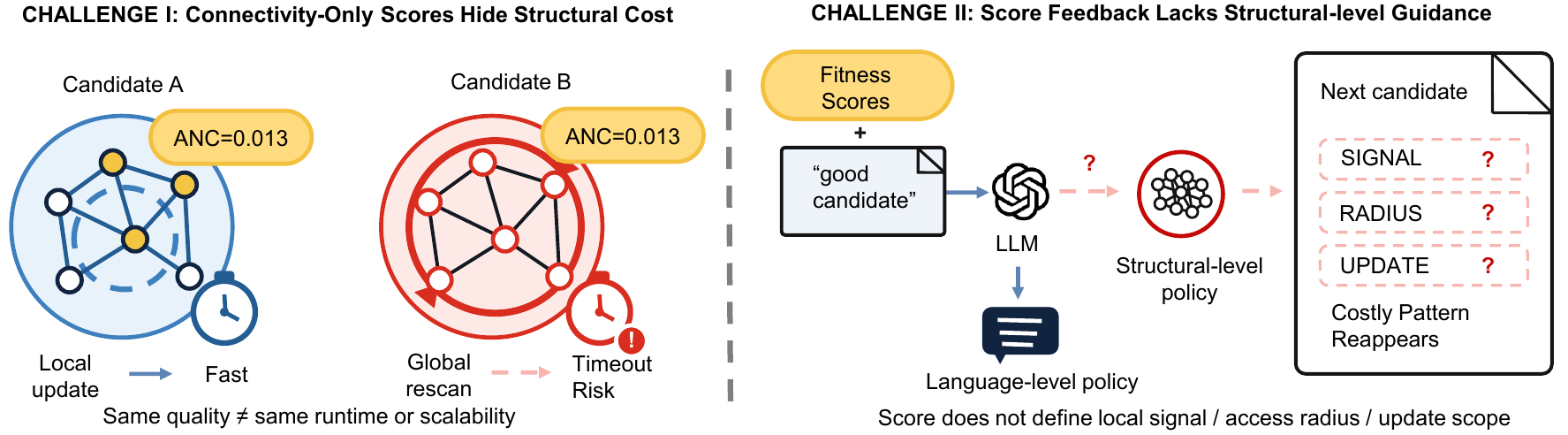}
 \caption{Two key challenges faced by LLM driven search for discovering network dismantling heuristics: exposing candidate runtime differences among program structures and converting execution feedback into structural guidance for subsequent code generation.}
 \Description{The figure illustrates the difficulty of revealing computational cost differences among program structures and of explicitly constraining subsequent code generation toward efficient local structures.}
 \label{fig:motivation}
\end{figure*}

However, LLM driven search for network dismantling still faces two challenges (Fig.~\ref{fig:motivation}). First, connectivity based fitness does not expose computational cost differences among program structures: an effective candidate may still depend on expensive whole graph computation and fail to scale. Second, scalar fitness or selection feedback~\cite{romeraparedes2024,zheng2025,chen2026} and language level reflection~\cite{ye2024,mao2024} can rank or revise candidates, but do not form executable structure level constraints. They do not specify which local signals to reuse or avoid, or how to bound neighborhood access and state updates, allowing costly patterns to recur in later programs.

To address both challenges, we propose AutoSND, a three stage framework that searches for complete dismantling programs built from local signal computation, residual-state maintenance, and bounded local updates. Stage I explores from a simple heuristic and records candidate quality, runtime, execution states, and code operations. Stage II applies LLM-Struct to each archived candidate, aligns execution outcomes with program structures, and compiles what to reuse, avoid, and bound into an explicit structural policy. Stage III conditions continued tree search on this policy and selects quality- and speed-prioritized candidates from the resulting Pareto frontier. We interpret the final programs not only as optimized dismantling heuristics but also as testable structural hypotheses, specified by their signals, neighborhood-access scopes, and state-update rules.

In summary, the contributions are as follows:
\begin{itemize}[leftmargin=*]
 \item We propose AutoSND, a three-stage automatic heuristic design framework that searches complete executable network dismantling heuristics rather than only node-scoring functions and jointly optimizes dismantling quality, candidate runtime, and executability through policy-guided tree search.

 \item We introduce Execution-to-Evidence Structural Policy Induction, which applies LLM-Struct to every archived candidate, aligns candidate outcomes, execution states, and code operations, and compiles them into explicit policies on local signals, neighborhood access, and update ranges that condition Stage III generation.

 \item We evaluate AutoSND on 12 real-world and 3 large real-world networks, demonstrating competitive dismantling quality, stable executability, and scalable, interpretable structures. Through structural-guidance ablations, search-lineage analysis, and code-component ablations, we derive reproducible principles for interpretable and efficient network dismantling heuristic design.
\end{itemize}

\section{Related Work}

\subsection{Network Dismantling Methods}

Network dismantling methods construct removal sequences from predefined structural signals or learned rules. Classical heuristics include residual degree based HDA, finite radius CI, Min Sum, BPD, residual 2 core based CoreHD, and cost aware generalized dismantling~\cite{holme2002,morone2015,braunstein2016,mugisha2016,zdeborova2016,ren2019}. Learning based methods include FINDER, GDM, NIRM, and symbolized reinforcement learning~\cite{fan2020,grassia2021,zhang2022,zheng2025symbolic}. Their signals and updates are manually specified, learned as fixed policies, or extracted post hoc.

\subsection{LLM Based Automatic Heuristic Design}

LLM driven AHD combines code generation, executable evaluation, and search. FunSearch, EoH, ReEvo, MCTS-AHD, and HiFo-Prompt organize generation through selection, joint idea and code evolution, reflection, tree planning, or historical insights~\cite{romeraparedes2024,liu2024eoh,ye2024,zheng2025,mao2024}. LLM4CN~\cite{zhang2025llm4cn} evolves static \texttt{score\_nodes(A)} functions from manually designed initial scores. Such methods use history mainly for selection or language feedback, not executable constraints specifying what structures to reuse, avoid, or bound.

\begin{figure*}[t]
 \centering
 \includegraphics[width=0.8\textwidth]{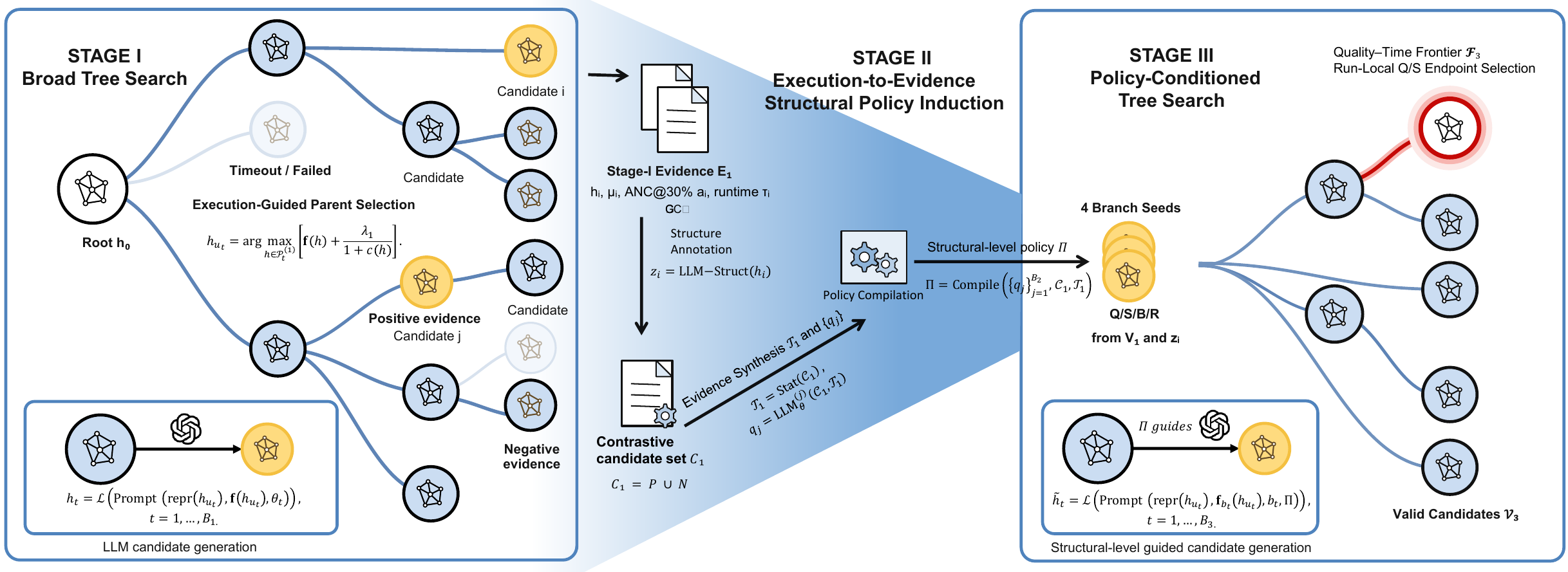}
 \caption{The three stage sequential workflow of AutoSND. AutoSND searches a network dismantling heuristic that can generate a complete removal sequence rather than a static scoring function that ranks original graph nodes once. Stage I generates and evaluates candidates from a simple heuristic and forms an execution evidence archive $E_1$. Stage II gives all candidates a unified structural description, constructs quality and candidate runtime contrastive evidence, and compiles historical statistics into a structural policy $\Pi$. Stage III separately selects quality, speed, balanced, and repair seeds from the valid Stage I candidate set $\mathcal V_1$ and continues multistep search conditioned on $\Pi$. The fixed evaluation framework is responsible for the task interface, search orchestration, unified evaluation, and validity verification, but it does not participate in candidate structure design.}
 \Description{The figure shows the sequential flow from Broad Tree Search through Execution-to-Evidence Structural Policy Induction to Policy-Conditioned Tree Search.}
 \label{fig:framework}
\end{figure*}

\section{Network Dismantling and AHD Definition}

\subsection{Network Dismantling Task}
\label{sec:nd-task}

Given a nonempty undirected graph $G=(V,E)$ with $N=|V|$ nodes, network dismantling requires outputting a complete removal sequence $\pi=(v_1,\ldots,v_N)$ such that network connectivity decreases as quickly as possible as nodes are removed~\cite{morone2015,braunstein2016}. Let
\begin{equation}
G_k=G\!\left[V\setminus\{v_1,\ldots,v_k\}\right],\quad
\sigma_{\mathrm{GCC}}(G)=\max_{C\in\mathcal{C}(G)}|C|,
\end{equation}
where $\mathcal{C}(G)$ is the set of connected components of $G$, and $\sigma_{\mathrm{GCC}}(\varnothing)=0$. Following the definition of accumulated normalized connectivity (ANC) in FINDER~\cite{fan2020}, AutoSND uses the largest connected component as the connectivity function:
\begin{equation}
\operatorname{ANC}(\mathrm{GCC},\pi)
=\frac{1}{N}\sum_{k=1}^{N}
\frac{\sigma_{\mathrm{GCC}}(G_k)}
{\sigma_{\mathrm{GCC}}(G)}.
\end{equation}
A lower ANC (GCC) indicates a faster decrease in network connectivity.

\subsection{Automatic Heuristic Design}

Given a heuristic space $\mathcal{H}$, a representative instance set $\mathcal{D}$, and a minimization objective $f$ for the original task, AHD searches for a heuristic with optimal expected performance~\cite{chen2026}:
\begin{equation}
h^\star=\arg\max_{h\in\mathcal{H}}\mathbf f(h),
\quad
\mathbf f(h)=\mathbb{E}_{x\in\mathcal{D}}
\left[-f\!\left(h(x)\right)\right].
\end{equation}
Here, $h^\star$ is the optimal heuristic algorithm, $x$ is a representative instance, $h(x)$ is the solution generated by the heuristic, and $\mathbf f(h)$ is the expected fitness of candidate $h$ on $\mathcal{D}$. The negative sign converts the minimization objective of the original task into fitness maximization. In LLM driven AHD, a new candidate is conditionally generated from the representation of its parent candidate and the current control information~\cite{chen2026}:
\begin{equation}
h_c=\mathcal{L}\!\left(
\operatorname{Prompt}\!\left(\operatorname{repr}(h_p),\theta^{(t)}\right)
\right),
\end{equation}
where $\mathcal{L}$ is the conditional LLM generator, $h_p$ is the parent, $\operatorname{repr}(h_p)$ is its description and code, $\operatorname{Prompt}$ constructs the prompt, $\theta^{(t)}$ is the round specific control information, and $h_c$ is the child. Here, $h$ implements a complete network dismantling heuristic rather than a single scoring function. It may jointly implement previously unspecified modules for structural signal computation, residual state maintenance, neighborhood access, node selection, and local updates, and output a complete removal sequence:
\begin{equation}
h:G\mapsto\pi_h=(v_1,\ldots,v_N).
\end{equation}

\section{The Three Stage AutoSND Framework for Automatic Heuristic Design}
\label{sec:method}

As shown in Fig.~\ref{fig:framework}, AutoSND treats a complete network dismantling heuristic as the search object and consists of \emph{Broad Tree Search} (Stage I), \emph{Execution-to-Evidence Structural Policy Induction} (Stage II), and \emph{Policy-Conditioned Tree Search} (Stage III). This process follows the generation, evaluation, and selection paradigm of LLM driven AHD~\cite{romeraparedes2024,ye2024,liu2024,zheng2025,chen2026}, and further transforms the relationships among code structure, dismantling quality, and candidate runtime obtained in Stage I into structural policies that directly condition subsequent search. The overall relationship among the three stages is
\begin{equation}
h_0
\xrightarrow{\mathcal S_1}E_1
\xrightarrow{\mathcal S_2:\,e_i\leftarrow(e_i,z_i)}
(E_1,\Pi)
\xrightarrow{\mathcal S_3}\mathcal V_3
\xrightarrow{\operatorname{Select}}(h_Q,h_S).
\end{equation}
Here, $h_0$ is the initial heuristic. $\mathcal S_1$ searches candidates from $h_0$ and forms execution evidence archive $E_1$. $\mathcal S_2$ applies LLM-Struct to every candidate, merges structural description $z_i$ into record $e_i$, and induces policy $\Pi$ from the updated archive. $\mathcal S_3$ selects four branch seed types from $\mathcal V_1$ and searches under $\Pi$. Candidates completing evaluation within the boundary of Stage $s$ form valid set $\mathcal V_s$, so $\mathcal V_1$ and $\mathcal V_3$ denote valid Stage I and III candidates. $h_Q$ and $h_S$ are the final quality and speed prioritized candidates.

\subsection{Broad Tree Search}
\label{sec:stage1}

Stage I uses HDA as a simple initial candidate. HDA removes a maximum degree node in the current residual graph and updates residual degrees after each removal~\cite{holme2002}. It provides only a low complexity root and does not require later candidates to retain one degree ordering: the LLM may change structural signals, neighborhood ranges, state maintenance, and local updates. To control online evaluation cost, Stages I and III evaluate the trajectory over the first 30\% of removed nodes using the normalized mean GCC over this interval:
\begin{equation}
\operatorname{ANC}@30\%(\mathrm{GCC},\pi)
=\frac{1}{K}\sum_{k=1}^{K}
\frac{\sigma_{\mathrm{GCC}}(G_k)}
{\sigma_{\mathrm{GCC}}(G)},
K=\operatorname{round}(0.30N).
\end{equation}
This quantity is the mean height of the normalized GCC trajectory over the first 30\% removal interval rather than the single point value at an exact removal ratio of 30\%. During search, candidates are compared only using ANC@30\% (GCC) and candidate runtime.

\paragraph{Candidate evaluation.}
For a valid candidate $h_i$, we jointly evaluate its connectivity $a(h_i)$ and candidate runtime $\tau(h_i)$ during search, denoted by $a(h_i)=a_i$ and $\tau(h_i)=\tau_i$:
\begin{equation}
a_i=\operatorname{ANC}@30\%(\mathrm{GCC},\pi_{h_i}),\quad
\tau_i=T_{\mathrm{end}}(h_i)-T_{\mathrm{start}}(h_i).
\end{equation}
To place quality and candidate runtime on a common scale, we rank-normalize the valid candidates generated so far. Let the current valid Stage I set be $\mathcal V_1$ and $M_1=|\mathcal V_1|$. When $M_1\geq2$, a lower-is-better metric $x$ is ranked in ascending order:
\begin{equation}
\rho(x_i)=1-\frac{\operatorname{rank}_{\mathrm{asc}}(x_i)-1}{M_1-1}.
\end{equation}
When $M_1=1$, the only rank-normalized value is 1. Here $N=|V|$ always denotes the input node count, whereas $M_1$ is the number of valid Stage I candidates. Candidate fitness is
\begin{equation}
\mathbf f(h_i)=
\alpha\,\rho\!\left(a_i\right)
+\beta\,\rho\!\left(\tau_i\right),
\end{equation}
where $\alpha$ and $\beta$ control the relative importance of dismantling quality and candidate runtime during search.

\paragraph{Candidate generation.}
At generation $i$, child $h_i$ is generated by providing the LLM with parent code $\operatorname{h}_{u_i}$, parent score $\mathbf f(h_{u_i})$, and exploration direction $\theta_i$:
\begin{equation}
h_i=
\mathcal L\!\left(
\operatorname{Prompt}
\left(
\operatorname{h}_{u_i},\mathbf f(h_{u_i}),\theta_i
\right)
\right).
\end{equation}

\paragraph{Parent candidate selection.}
For $s\in\{1,3\}$, let $\mathcal P_t^{(s)}$ be the available parent pool at expansion $t$ of the corresponding search stage, and let $c(h)$ be the number of children already produced by $h$. Parent priority is
\begin{equation}
U_s(h)=\mathbf f(h)+\frac{\lambda_s}{1+c(h)},
\quad
h_{u_t}=\arg\max_{h\in\mathcal P_t^{(s)}}U_s(h).
\end{equation}
The first term deepens paths with better quality or efficiency, whereas the second gives less explored paths more opportunities, preventing premature concentration on one branch.

\paragraph{Forming the execution evidence archive.}
After Stage I, AutoSND archives the search tree and all candidate records from this stage to obtain execution evidence archive $E_1$:
\begin{equation}
E_1=
\left(
\mathcal G_1,\{e_i\}_{i=1}^{N_1}
\right),
\quad
e_i=(h_i,\mu_i,a_i,\tau_i).
\end{equation}
Here, $\mathcal G_1$ stores the search tree, $h_i$ is the $i$th candidate, and $\mu_i$ records generation context and execution state such as success, timeout, or runtime error. For $h_i\in\mathcal V_1$, $a_i$ and $\tau_i$ record ANC@30\% (GCC) and candidate runtime. For other candidates, $a_i$ is empty, candidate runtime is truncated on timeout, and $\mu_i$ identifies the state. Every candidate is retained in $E_1$, but only $\mathcal V_1$ enters quality and candidate runtime ranking.

\subsection{Execution-to-Evidence Structural Policy Induction}
\label{sec:stage2}

Stage I identifies which candidates perform better but not which structures subsequent search should reuse. Stage II takes $E_1$ as historical input, applies LLM-Struct to every candidate, aligns candidate structures with quality, candidate runtime, and execution states, and induces policy $\Pi$. It generates no new candidate and does not change Stage I evaluation results.

\subsubsection{Unified Structural Description and Contrastive Candidates}

First, the LLM performs a unified structural analysis of all candidates in $E_1$:
\begin{equation}
z_i=\operatorname{LLM-Struct}(h_i),\quad i=1,\ldots,N_1.
\end{equation}
$z_i$ summarizes the candidate's overall structural family, local signals, neighborhood access scope, state update mode, and high cost structures. After normalization, it is merged with the corresponding execution record $e_i$:
\begin{equation}
E_1\leftarrow\left(\mathcal G_1,\{(e_i,z_i)\}_{i=1}^{N_1}\right).
\end{equation}

Contrastive candidates are then selected from the updated $E_1$ by quality, candidate runtime, and execution status, forming positive and negative evidence:
\begin{equation}
\mathcal C_1=\mathcal P\cup\mathcal N,\quad
\mathcal N=\mathcal N^{a}\cup\mathcal N^{\tau}\cup\mathcal N^{\mathrm{fail}}.
\end{equation}
The positive set $\mathcal P$ is selected from leading Pareto layers of $(a_i,\tau_i)$. The disjoint negative set $\mathcal N$ is partitioned by primary disadvantage into poor quality $\mathcal N^{a}$, high candidate runtime $\mathcal N^{\tau}$, and invalid evaluation $\mathcal N^{\mathrm{fail}}$. A candidate disadvantaged in both quality and candidate runtime is assigned to only its primary class. Appendix~\ref{app:implementation} gives the matching rules.

\subsubsection{Historical Statistics and Formation of the Structural Policy}

Stage II compares each sufficiently supported structural item or low order combination between positive and negative candidates. It calculates occurrence proportions in $\mathcal P$ and $\mathcal N$ and, for valid candidates, summarizes $a_i$, $\tau_i$, and neighborhood access and state update scopes explicitly specified or parsable in code. These results are
\begin{equation}
\mathcal T_1=\operatorname{Stat}(\mathcal C_1).
\end{equation}
$\mathcal T_1$ uses $\mathcal N$ as a unified negative contrast while retaining separate $\mathcal N^{a}$, $\mathcal N^{\tau}$, and $\mathcal N^{\mathrm{fail}}$ results. It therefore distinguishes structures associated mainly with quality degradation, slow execution, or failure and records structural frequencies and failure risks. The LLM reads both $\mathcal C_1$ and $\mathcal T_1$ and proposes structural rules:
\begin{equation}
q_j=\operatorname{LLM}_{\theta}^{(j)}(\mathcal C_1,\mathcal T_1),
\quad j=1,\ldots,B_2.
\end{equation}
Here, $B_2$ is the number of Stage II induction requests. Each $q_j$ proposes which structures to reuse or avoid and how to bound access and update scopes. It generates no candidate. All proposals are compiled against the same historical evidence:
\begin{equation}
\Pi=
\operatorname{Compile}\!\left(
\{q_j\}_{j=1}^{B_2},\mathcal C_1,\mathcal T_1
\right)
=\left(\mathcal A^{+},\mathcal A^{-},\mathcal B\right).
\end{equation}
$\mathcal A^{+}$ records structural families, local signals, and update modes to reuse. $\mathcal A^{-}$ records structures and costly patterns to avoid. $\mathcal B$ gives recommended bounds for neighborhood access and state updates. These parts answer what to reuse, what to avoid, and how far computation should extend, making $\Pi$ an explicit ``reuse, avoid, and bound'' constraint. Section~\ref{sec:setup} gives statistical requirements. Appendices~\ref{app:implementation} and~\ref{app:evidence} give the structural vocabulary and source tracing rules.

\subsection{Policy-Conditioned Tree Search}
\label{sec:stage3}

\subsubsection{Four Types of Branch Seed Selection}

Stage III selects seeds from the valid Stage I candidate set $\mathcal V_1$. Failed candidates are excluded as parents and affect Stage III only through the negative evidence compiled into $\Pi$. To remove generation order effects, AutoSND recomputes the quality and candidate runtime ranks on $\mathcal V_1$ and uses the common branch score
\begin{equation}
\mathbf f_b(h_i)=\alpha_b\rho(a_i)+\beta_b\rho(\tau_i),
\alpha_b+\beta_b=1, b\in\{\mathrm Q,\mathrm S,\mathrm B,\mathrm R\}.
\end{equation}
The four branches retain complementary roles. Q emphasizes quality. S applies a quality gate before prioritizing candidate runtime. B combines quality and candidate runtime with round robin sampling across structural families. R starts from quality promising or relatively slow candidates to seek lower candidate runtime.
These rules form seed sets $\mathcal H_{\mathrm Q}$, $\mathcal H_{\mathrm S}$, $\mathcal H_{\mathrm B}$, and $\mathcal H_{\mathrm R}$ for the four Stage III branches, with $b_t\in\{\mathrm Q,\mathrm S,\mathrm B,\mathrm R\}$ denoting the branch used at expansion $t$. The complete set definitions, thresholds, weights, quotas, and overlap handling are given in Appendix~\ref{app:search}.

\begin{table*}[t]
  \caption{Unified comparison of native, AHD, AI-agent, and AutoSND methods on the 12 main graphs. Graph and Mean columns report complete-sequence $100\times\operatorname{ANC}(\mathrm{GCC})$; lower is better. Means and average ranks are recomputed from the full-precision values over the 18 methods completing all 12 graphs. \texttt{TO} means that no valid complete sequence was returned within 600~s, while ``--'' indicates an undefined summary or candidate validity. Coverage is the number of completed main graphs, and mean time is computed over completed inputs. Best and second-best values in the per-graph quality, Mean, and Avg.\ rank columns are shown in red boldface and underlined, respectively. Full-precision results, baseline reevaluation scope, and timing sources are given in Appendices~\ref{app:data} and~\ref{app:fullresults}.}
  \label{tab:maincomparison}
  \centering
  \fontsize{6pt}{6pt}\selectfont
  \setlength{\tabcolsep}{2.0pt}
  \renewcommand{\arraystretch}{0.86}
  \begin{adjustbox}{width=\textwidth}
  \begin{tabular}{l*{12}{r}rrrrr}
    \toprule
    Methods & CEnew & Collab. & condmat & crime & email & Grid &
    hamster & HepPh & Yeast & HI-II-14 & Digg & Gnutella31 &
    Mean & Avg.\ rank & Valid & Cov. & Time (s) \\
    \midrule
    \rowcolor[RGB]{226,239,248}
    \multicolumn{18}{c}{\textbf{Native Methods}} \\
    HDA & 10.89 & 10.71 & 13.03 & 11.33 & 22.54 & 5.51 & 18.69 & 18.47 & 13.89 & 5.78 & 8.81 & 11.46 & 12.59 & 11.17 & -- & 12/12 & 0.35 \\
    CoreHD & 10.05 & 10.76 & 13.06 & 11.65 & 22.58 & 5.06 & 19.00 & 18.47 & 14.01 & 5.71 & 8.62 & 11.29 & 12.52 & 10.83 & -- & 12/12 & 0.34 \\
    DC & 11.82 & 13.16 & 14.73 & 12.39 & 25.06 & 6.52 & 20.01 & 23.00 & 19.58 & 6.94 & 9.67 & 12.94 & 14.65 & 14.83 & -- & 12/12 & 0.11 \\
    CI & 12.87 & 13.68 & 19.39 & 15.22 & 27.06 & 7.95 & 21.00 & 26.87 & 23.69 & 8.93 & 10.85 & 14.47 & 16.83 & 15.92 & -- & 12/12 & 0.79 \\
    KCore & 18.52 & 22.82 & 27.40 & 26.02 & 29.48 & 26.31 & 25.73 & 28.32 & 28.93 & 10.84 & 10.97 & 14.35 & 22.47 & 16.75 & -- & 12/12 & 0.31 \\
    CLUC & 49.42 & 46.01 & 48.68 & 46.26 & 42.77 & 29.45 & 48.34 & 47.84 & 39.76 & 15.50 & 13.16 & 15.56 & 36.90 & 18.00 & -- & 12/12 & 0.79 \\
    GND & \underline{9.04} & 10.70 & 12.65 & 11.96 & 22.31 & 4.12 & \textcolor{red}{\textbf{14.25}} & 18.42 & 12.09 & 6.06 & 8.72 & 12.05 & 11.86 & 8.08 & -- & 12/12 & 14.55 \\
    MinSum & 9.25 & 9.58 & 12.33 & 10.94 & 22.14 & 4.36 & 17.43 & 17.24 & 13.65 & \underline{5.55} & \textcolor{red}{\textbf{8.55}} & 11.38 & 11.87 & 4.67 & -- & 12/12 & 14.04 \\
    BPD & 9.28 & 10.77 & 12.76 & 11.07 & 22.36 & 4.42 & 18.06 & 17.70 & 14.31 & 5.92 & 8.75 & 12.16 & 12.30 & 10.00 & -- & 12/12 & 11.76 \\
    FINDER & 10.70 & 11.43 & 13.02 & 10.99 & 22.27 & 5.00 & 18.92 & 19.82 & 13.46 & \textcolor{red}{\textbf{5.54}} & 8.66 & \textcolor{red}{\textbf{11.10}} & 12.58 & 8.75 & -- & 12/12 & 6.95 \\
    % \midrule
    \rowcolor[RGB]{226,239,248}
    \multicolumn{18}{c}{\textbf{LLM AHD Methods}} \\
    FunSearch & 26.71 & 9.14 & 11.07 & 11.03 & \textcolor{red}{\textbf{21.12}} & \textcolor{red}{\textbf{3.49}} & 16.75 & 16.53 & 12.59 & 11.29 & 8.70 & 11.61 & 13.34 & 6.75 & 98.4\% & 12/12 & 79.23 \\
    Clade-AHD & 9.68 & \underline{7.54} & 11.48 & 11.04 & \underline{21.20} & 3.89 & 15.28 & 14.49 & 13.73 & 5.59 & 8.60 & 11.78 & 11.19 & 4.92 & 72.2\% & 12/12 & 93.91 \\
    ERA & 9.48 & 8.52 & \underline{10.81} & 11.05 & 21.57 & 4.02 & \underline{14.86} & 15.15 & \underline{12.07} & 5.65 & 8.58 & 11.65 & 11.12 & \underline{3.92} & 79.6\% & 12/12 & 29.29 \\
    MCTS-AHD & 9.69 & 10.18 & 12.45 & 11.16 & 22.09 & 4.44 & 17.49 & 17.87 & 13.59 & 5.69 & 8.62 & 11.77 & 12.09 & 7.58 & 75.6\% & 12/12 & 94.28 \\
    AlphaEvolve & 9.88 & 10.71 & 13.01 & 11.12 & 22.31 & 4.68 & 18.92 & 18.58 & 14.30 & 5.79 & 8.66 & 11.96 & 12.49 & 10.50 & 89.4\% & 12/12 & 9.01 \\
    LLM4CN & 12.50 & 8.72 & \texttt{TO} & 12.43 & 23.98 & 5.93 & 16.84 & 14.73 & 13.25 & 6.34 & \texttt{TO} & \texttt{TO} & -- & -- & 93.8\% & 9/12 & 76.37 \\
    EoH & 10.01 & 10.73 & 13.00 & 11.14 & 22.25 & 4.58 & 18.80 & 18.70 & 14.19 & 5.75 & 8.61 & 11.80 & 12.46 & 9.83 & 95.4\% & 12/12 & 58.06 \\
    ReEvo & 9.92 & 10.02 & 12.49 & 11.29 & 22.38 & 4.85 & 18.18 & \texttt{TO} & 13.63 & 6.00 & 8.77 & 12.29 & -- & -- & 60.9\% & 11/12 & 23.31 \\
    HiFo-Prompt & 10.02 & 11.53 & 13.73 & 11.18 & 22.64 & 4.98 & 19.98 & \texttt{TO} & 14.87 & 6.00 & 8.75 & 12.33 & -- & -- & 100.0\% & 11/12 & 16.54 \\
    HSEvo & 9.42 & 10.34 & 11.75 & 11.16 & 21.88 & 4.32 & 16.92 & \texttt{TO} & 15.91 & 5.71 & \texttt{TO} & \texttt{TO} & -- & -- & 67.7\% & 9/12 & 47.42 \\
    % \midrule
    \rowcolor[RGB]{226,239,248}
    \multicolumn{18}{c}{\textbf{AI agent Methods}} \\
    Codex-Evo-Q & 9.68 & 10.05 & \texttt{TO} & \textcolor{red}{\textbf{10.71}} & 21.87 & 4.40 & 17.39 & \texttt{TO} & 13.34 & 5.60 & \texttt{TO} & \texttt{TO} & -- & -- & 23.6\% & 8/12 & 15.25 \\
    Codex-Direct-Q & \textcolor{red}{\textbf{8.93}} & \texttt{TO} & \texttt{TO} & \underline{10.82} & 21.51 & \texttt{TO} & \texttt{TO} & \texttt{TO} & \texttt{TO} & \texttt{TO} & \texttt{TO} & \texttt{TO} & -- & -- & 60.0\% & 3/12 & 71.68 \\
    \midrule
    \rowcolor[RGB]{226,239,248}
    \multicolumn{18}{c}{\textbf{Ours}} \\
    \textbf{AutoSND-S} & 9.80 & \textcolor{red}{\textbf{7.43}} & 10.83 & 11.29 & 21.70 & 4.58 & 14.99 & \textcolor{red}{\textbf{13.67}} & \textcolor{red}{\textbf{11.42}} & 5.57 & 8.73 & \underline{11.14} & \textcolor{red}{\textbf{10.93}} & 5.00 & 99.0\% & 12/12 & 2.06 \\
    \textbf{AutoSND-Q} & 9.66 & 7.71 & \textcolor{red}{\textbf{10.77}} & 10.99 & 21.67 & \underline{3.87} & 15.19 & \underline{14.11} & 12.10 & 5.57 & \underline{8.57} & 11.66 & \underline{10.99} & \textcolor{red}{\textbf{3.50}} & 99.0\% & 12/12 & 2.86 \\
    \bottomrule
  \end{tabular}
  \end{adjustbox}
\end{table*}

\subsubsection{Three-Stage Tree Search and Final Candidate Selection}

After the four branches are initialized with $\mathcal H_{\mathrm Q}$, $\mathcal H_{\mathrm S}$, $\mathcal H_{\mathrm B}$, and $\mathcal H_{\mathrm R}$, Stage III expands them in parallel. Each generation is conditioned on the parent candidate code, branch score, branch role, and structural policy:
\begin{equation}
\widetilde h_t=
\mathcal L\!\left(
\operatorname{Prompt}\!\left(
\operatorname{repr}(h_{u_t}),
\mathbf f_{b_t}(h_{u_t}),
b_t,\Pi
\right)
\right), t=1,\ldots,B_3.
\end{equation}
where $B_3$ is the Stage III candidate budget. Each branch uses the $U_3$ defined in Section~\ref{sec:stage1} and selects parents from its valid tree nodes. A valid child can subsequently serve as a parent, enabling multistep tree deepening rather than single-pass variant generation from fixed seeds. The remaining generation, evaluation, and archiving procedures follow Stage I.

After Stage III, let $\mathcal V_3$ denote its valid candidate set. A candidate $h_i$ dominates $h_j$ if $a(h_i)\leq a(h_j)$ and $\tau(h_i)\leq\tau(h_j)$, with at least one strict inequality. The nondominated candidates form the quality--candidate-runtime frontier $\mathcal F_3$. Its quality endpoint is selected lexicographically by
\begin{equation}
h_Q^{\mathrm{run}}=
\arg\min_{h\in\mathcal F_3}
\left(a(h),\tau(h)\right).
\end{equation}

To obtain a quality and speed-prioritized endpoint, we use:
\begin{equation}
\mathcal F_3^{\mathrm S}
:=
\left\{
h\in\mathcal F_3:
\rho_{\mathcal F_3}(a(h))\geq r_{\mathrm S}
\right\},
\end{equation}
\begin{equation}
h_S^{\mathrm{run}}
:=
\arg\min_{h\in\mathcal F_3^{\mathrm S}}
\left(\tau(h),a(h)\right).
\end{equation}
where $\rho_{\mathcal F_3}(a(h))\in[0,1]$ is the normalized quality rank recomputed on $\mathcal F_3$ using the normalization in Eq.~(9), with 1 indicating the best ANC@30\% (GCC). Among the candidates satisfying the fixed quality threshold $r_{\mathrm S}=0.55$, AutoSND selects the candidate with the lowest runtime; ANC@30\% (GCC) and archive order break exact ties. The complete records are provided in Appendices~\ref{app:search} and~\ref{app:evidence}.

\section{Experiments}
\label{sec:experiments}

This section addresses three questions: RQ1, whether AutoSND can discover network dismantling heuristics that have competitive quality and can be executed stably, RQ2, whether the structural policy induced by Stage II can guide Stage III to expand the candidate region and deepen effective lineages, and RQ3, what code structures are formed by the final candidates that can explain their quality and efficiency. We denote the frozen quality side and speed side representatives by \textbf{AutoSND-Q} ($h_Q$) and \textbf{AutoSND-S} ($h_S$), respectively, and test the quality, efficiency, and scalability of the two fixed candidates on external networks. Candidate identities and generation records are given in Appendix~\ref{app:evidence}.

\begin{figure*}[t]
\centering
\includegraphics[width=0.8\textwidth]{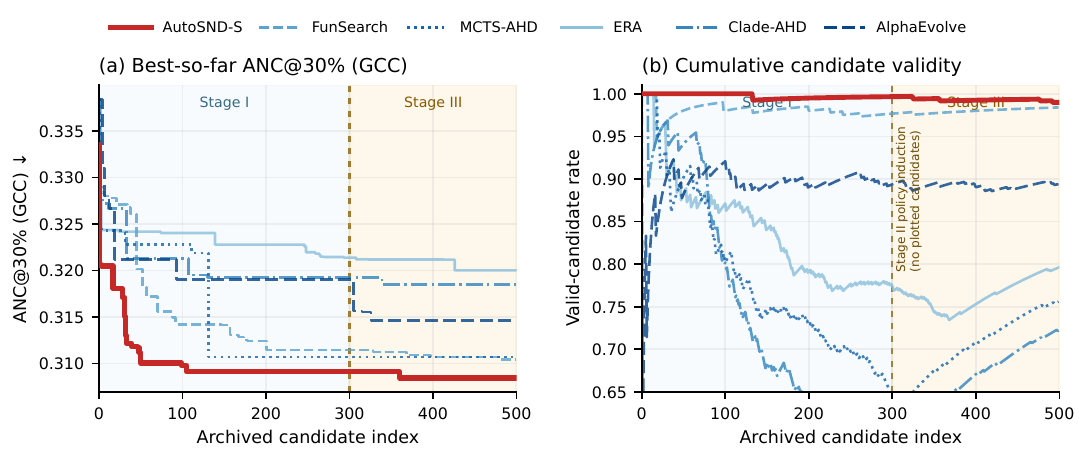}
\caption{Stagewise search progress of AutoSND-S and AHD methods. The left panel shows the best observed ANC@30\% (GCC), and the right panel shows cumulative archive validity. The red solid line denotes AutoSND-S, and the vertical dashed line separates Stage I and Stage III.}
\label{fig:trajectory}
\end{figure*}

\subsection{Experimental Setup}
\label{sec:setup}

\subsubsection{Datasets}

The main evaluation uses 12 real networks, and  spans power-grid infrastructure (Grid), protein interaction (Yeast), scientific-collaboration networks (Collaboration, condmat, and HepPh), communication and social systems (email, hamster, and Digg), and so on, spanning 453--62,561 nodes and 1,473--147,878 edges. Large scale evaluation uses Facebook, YouTube, and Flickr from FINDER~\cite{fan2020}, reaching 1,624,991 nodes and 15,473,043 edges. This tests whether the discovered pattern transfers across distinct network formation regimes rather than merely increasing benchmark count. To improve efficiency, only a 500 node powerlaw network is used in the search phase, while also avoiding test set leakage. Data and preprocessing are detailed in Appendix~\ref{app:data} and Appendix~\ref{app:scaling}.

\subsubsection{Evaluation Metrics}
\label{sec:evaluation-metrics}

\paragraph{Quality and candidate runtime.}
Complete sequence quality is $100\times\operatorname{ANC}(\mathrm{GCC},\pi)$ as defined in Section~\ref{sec:nd-task}. Means and average ranks include only methods completing all 12 main graphs. Completeness requires a valid full sequence within 600 s. Candidate runtime covers only sequence generation by the frozen candidate, and is averaged over completed inputs from the same implementation. Full per-graph results and high-precision summaries are provided in Appendices ~\ref{app:data}.

\paragraph{Reliability.}
Candidate validity is the fraction of generated candidates passing the unified checks, whereas complete sequence coverage is the fraction of test graphs completed by the Q/S candidates. AutoSND computes validity over Stages I/III. \texttt{TO}, invalid, and missing cases are not imputed. Formulas and budget accounting are in Appendix~\ref{app:search}.

\subsubsection{Baseline Methods}

Native baselines are HDA~\cite{holme2002}, CoreHD~\cite{zdeborova2016}, DC, CI~\cite{morone2015}, GND~\cite{grassia2021}, MinSum~\cite{braunstein2016}, BPD~\cite{mugisha2016}, KCore, and CLUC. AHD baselines are local reproductions of FunSearch~\cite{romeraparedes2024}, ReEvo~\cite{ye2024}, HSEvo~\cite{liu2024}, AlphaEvolve~\cite{novikov2025}, EoH~\cite{liu2024eoh}, MCTS-AHD~\cite{zheng2025}, Clade-AHD~\cite{chen2026}, HiFo-Prompt~\cite{mao2024}, ERA~\cite{zheng2025era}, and LLM4CN~\cite{zhang2025llm4cn} under the unified dismantling interface. Codex-Direct and Codex-Evo are equal-budget AI agent controls for independent generation and simplified evolutionary search under the same interface, proxy, and evaluator. Reproduction and control details are in Appendix~\ref{app:data}.

\subsubsection{Search Settings}
\label{sec:search-settings}

Main AutoSND and AHD searches use gpt-5.5, reasoning effort \texttt{none}, and temperature 0.2. AutoSND budgets are $(B_1,B_2,B_3)=(300,10,200)$, with Stages I/III counting candidates and Stage II counting induction calls. AutoSND-Q/S are frozen from the same complete full-AutoSND \texttt{Powerlaw\_500} source search with search seed 42, whereas each ablation's Q/S candidates are frozen from its corresponding complete ablation search with search seed 42 before external evaluation. Full configurations and provenance are in Appendices~\ref{app:search} and~\ref{app:evidence}.

\subsection{RQ1: Can AutoSND Reliably Discover Competitive and Low Structural Cost Heuristics?}

\subsubsection{Quality, Efficiency, and Coverage on Real Networks}

As shown in Table~\ref{tab:maincomparison}, the unified protocol in Section~\ref{sec:setup} evaluates complete-sequence quality, candidate validity, coverage, and candidate runtime. Among the 18 methods completing all 12 graphs, AutoSND-Q/S obtain the two lowest mean $100\times\operatorname{ANC}(\mathrm{GCC})$ values, 10.99 and 10.93, respectively. AutoSND-Q ranks first on average at 3.50, while AutoSND-S averages 5.00. The two candidates rank first on four graphs, and at least one ranks among the top two on seven. Both cover 12/12 graphs with 99.00\% candidate validity and mean candidate runtimes of 2.86 and 2.06 s, respectively. In contrast, Codex-Direct/Evo cover only 3/12 and 8/12 graphs, and Codex-Evo yields 23.60\% valid candidates. Thus, under the fixed interface, budget, and evaluator, AutoSND combines competitive dismantling quality, complete coverage, and low execution cost, while structured evidence accumulation and multistage search produce more reliable candidates than the two Codex controls.

\subsubsection{Advancing the Quality Frontier with Stable Candidate Generation}

This experiment compares AutoSND-S with five AHD methods using complete search seed 42 records. As shown in Fig.~\ref{fig:trajectory}, the best ANC@30\% (GCC) observed up to each candidate and cumulative archive validity track both the quality frontier and executable candidate supply. AutoSND-S improves rapidly early in the search and continues improving in Stage III. Its cumulative validity remains near 100\% and ends at 99.0\%. Together, the curves show that the three stage search advances the quality frontier while maintaining stable candidate generation.

\begin{figure*}[t]
\centering
\includegraphics[width=0.8\textwidth]{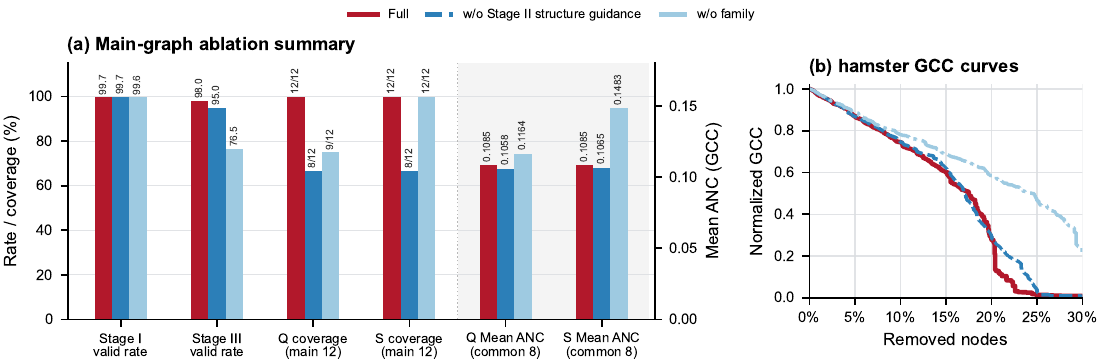}
\caption{Structural-guidance ablation of AutoSND-Q/S. Left: Stage I/III validity, Q/S main-12 coverage, and mean ANC (GCC) on eight common graphs (CEnew, Collaboration, crime, email, Grid, hamster, Yeast, HI-II-14). Colors denote \texttt{Full}/\texttt{w/o Stage II}/\texttt{w/o family}; exact labels use the left percentage axis for the first four groups and the right ANC (GCC) axis for the last two. Right: normalized GCC of final S candidates on hamster through 30\% removal.}
\label{fig:ablation}
\end{figure*}

\begin{figure}[t]
\centering
\includegraphics[width=0.8\columnwidth]{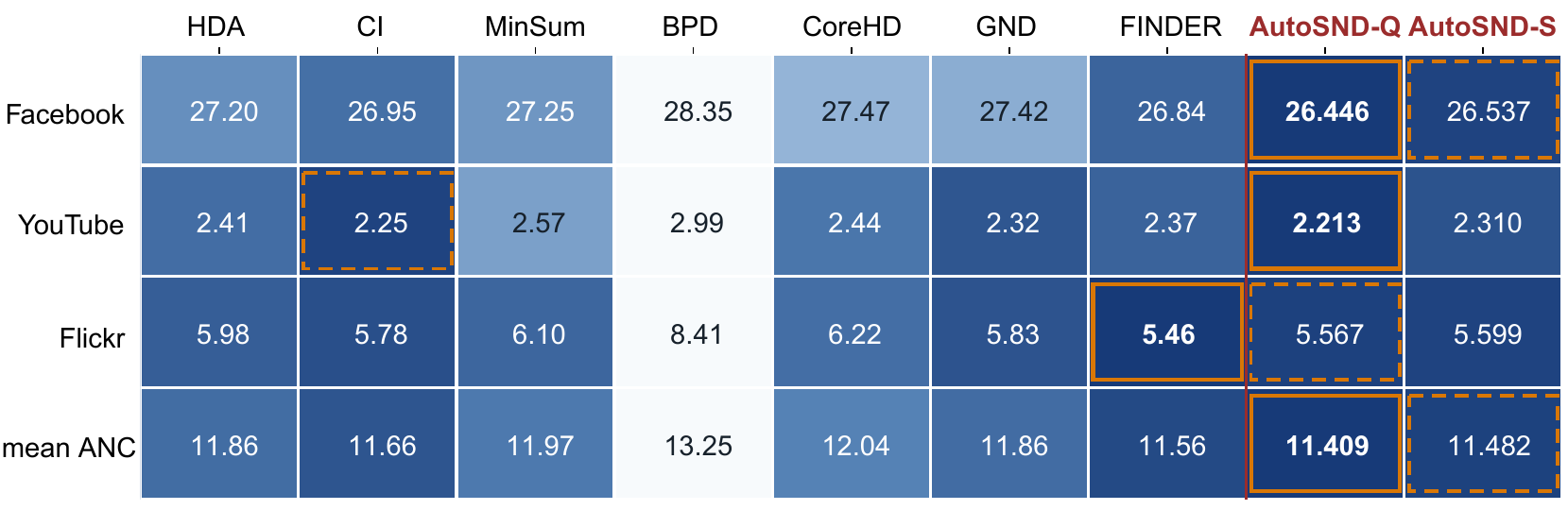}
\caption{Complete sequence quality on three large real networks. The first three rows report for each graph $100\times\operatorname{ANC}(\mathrm{GCC})$, and the last row reports the three graph mean. Lower is better. Baseline values are from FINDER~\cite{fan2020}, and AutoSND-Q/S values are from strict complete sequence recomputation. The complete sequence checks and released curve protocol are given in Appendix~\ref{app:scaling}.}
\label{fig:large}
\end{figure}

\begin{figure*}[t]
\centering
\includegraphics[width=0.8\textwidth]{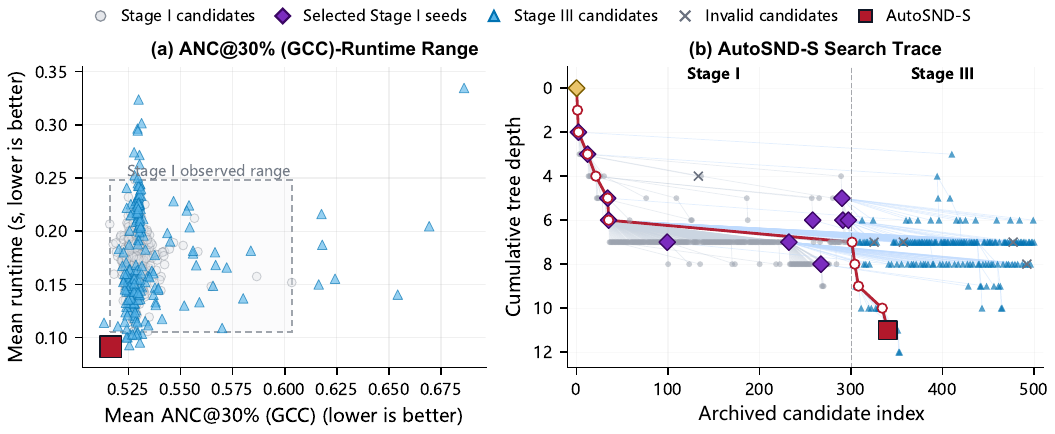}

\caption{Stagewise candidate distribution and AutoSND-S lineage on \texttt{Powerlaw\_500}. (a) The 500 candidates archived by source search seed 42 are positioned by mean ANC@30\% (GCC) and mean candidate runtime after reevaluation on proxy graph seeds 42--44. The dashed rectangle bounds the valid Stage I candidates. (b) The complete source-search tree retains all 300 Stage I and 200 Stage III candidates and their parent-to-child relationships. The red path and square mark the AutoSND-S lineage.}
\label{fig:lineage}
\end{figure*}

\subsubsection{Quality and Scalability on Very Large Real Networks}
We test three increasingly large social networks: Facebook, with 63,392 nodes and 816,831 edges, YouTube, with 1,134,890 nodes and 2,987,624 edges, and Flickr, with 1,624,991 nodes and 15,473,043 edges. As shown in Fig.~\ref{fig:large}, AutoSND-Q obtains the lowest values on Facebook and YouTube, and its mean ANC (GCC) over the three graphs is 11.409, below FINDER's 11.555. AutoSND-S averages 11.482 and also outperforms FINDER. Both candidates return valid complete sequences on all three networks, showing that their bounded local decisions and state maintenance scale to million-node networks while retaining competitive complete-sequence quality.

\subsection{RQ2: Can the Structural Policy Induced by Stage II Effectively Guide Stage III Search?}

\subsubsection{Ablation Experiments of AutoSND}
\label{sec:ablation}

As shown in Fig.~\ref{fig:ablation}, under the unified protocol in Section~\ref{sec:search-settings}, full AutoSND is compared with two ablations: \texttt{w/o Stage II structure guidance} bypasses policy induction, whereas \texttt{w/o structural family restriction} retains the policy but selects 24 Stage III seeds by one quality ranking. Both ablations follow Stage I, whose validity is thus stable at 99.7\%, 99.7\%, and 99.6\% for \texttt{Full}, \texttt{w/o Stage II}, and \texttt{w/o family}. Full AutoSND reaches 98.0\% Stage III validity. Q/S cover 12/12 graphs and average 0.1085 ANC (GCC) on the common eight. Without Stage II, both cover only 8/12 despite slightly lower common-graph ANC (GCC), locating the main loss in cross-network execution stability. Without family restriction, Stage III validity falls to 76.5\%. Q/S cover 9/12 and 12/12, and ANC (GCC) rises to 0.1164/0.1483, 7.3\%/36.7\% higher. On hamster, full AutoSND fragments the graph faster after about 20\% removal, the no-Stage-II variant reaches a similar level later, and the no-family-restriction variant is slowest. Thus, the explicit policy primarily supports cross-network stability, while structure-diverse seed selection limits invalid exploration and preserves quality.

\subsubsection{Stage III Expands the Search Region and Deepens Effective Lineages under Structural Guidance}

Figure~\ref{fig:lineage} examines whether policy-conditioned Stage III merely refines the Stage I population or reaches new candidate regions. In panel (a), 13.8\% of the 196 valid Stage III candidates fall outside the Stage I quality--candidate-runtime range, and 5.6\% cross at least one better boundary, showing that Stage III does not remain confined to the Stage I search region. In panel (b), the Stage I seed anchoring the AutoSND-S lineage receives 58.5\% of the subsequent expansions and produces 13 of the 16 descendants that improve both metrics. This lineage extends through five additional Stage III levels and reaches cumulative depth 11. Relative to their Stage I seeds, AutoSND-Q/S reduce ANC@30\% (GCC) by 1.9\%/1.6\% and candidate runtime by 24.2\%/39.0\%.

\subsection{RQ3: What Interpretable Structures Are Formed by AutoSND-Q/S?}

\subsubsection{Code Structure Analysis of AutoSND-Q/S}

\begin{table}[h]
\caption{Code structure comparison of AutoSND-Q/S.}
\label{tab:code}
\centering
\scriptsize
\resizebox{\columnwidth}{!}{%
\begin{tabular}{p{0.20\columnwidth}p{0.36\columnwidth}p{0.36\columnwidth}}
\toprule
Code structure & AutoSND-Q & AutoSND-S\\
\midrule
State representation & Static adjacency cache, active node set, residual degree & Mutable residual graph, residual degree\\
Main term & Dynamic residual degree & Compressed degree term $\log(1+d)\sqrt d$\\
Local correction & Frontier, boundary, weak connection, bridge, bounded two-hop, redundancy & Frontier/reachability, boundary ratio, weak connection, bridge, low degree pressure, redundancy\\
Neighbor access cap & 12, 16, and 24 in three stages & Increases by stage to 64\\
Two-hop access cap & 6, 8, and 8 in three stages & Bounded further access for each neighbor\\
Refresh after removal & All direct neighbors and at most 18 additional two-hop nodes & Direct neighbors, bounded two-hop nodes, and total affected set\\
\bottomrule
\end{tabular}}
\end{table}

We compare the structures of the two final candidates and uses the component removal experiment in Appendix~\ref{app:scaling} to check whether the main structures contribute to complete sequence quality. As shown in Table~\ref{tab:code}, their state representations, ranking terms, and update scopes are summarized. The two candidates can be summarized as
\begin{equation}
\operatorname{score}(v)
=\phi(d_v)+\psi_{\mathrm{local}}(v)-\omega_{\mathrm{red}}(v).
\end{equation}
where $\phi(d_v)$ is the residual degree backbone based on the current degree $d_v$, $\psi_{\mathrm{local}}(v)$ is a bounded local signal that captures fragmentation effects around node $v$, and $\omega_{\mathrm{red}}(v)$ measures local structural redundancy.
Q emphasizes boundary correction, whereas S emphasizes compressed degree and bounded state maintenance. Both use a lazy heap and local refresh. In Appendix~\ref{app:scaling}, complete Q/S reduce mean ANC (GCC) by 12.4\%/12.9\% relative to residual degree alone. Removing redundancy, weak connection/bridge, or boundary terms also increases ANC (GCC). Performance therefore comes from the backbone plus bounded local corrections.

\subsubsection{Implications for Heuristic Design}

AutoSND-Q/S suggest a testable principle: use incrementally maintainable residual degree as the backbone, bounded boundary/weak-connection/redundancy signals to adjust node order, and restricted access and updates to control maintenance cost. Component removal in Appendix~\ref{app:scaling} confirms contributions from the backbone and corrections, while their implementations are identified, as shown in Table~\ref{tab:code}. AutoSND-S reveals a more specific mechanism behind this pattern. At 14\% removal, its program reduces the hub coefficient while increasing the frontier, boundary, weak-tie, and bridge coefficients, the penalty on redundant neighborhoods, and the bounded neighbor-sampling cap. Residual degree therefore remains the maintained state backbone but no longer determines the next removal by itself. On hamster, at 20.35\% and 20.40\% removal, AutoSND-S selects nodes 170 and 1369 with residual degrees only 6 and 7, although hundreds of remaining nodes have higher degree. Removing node 170 splits the current 527-node GCC into components of 418 and 108 nodes; removing node 1369 then splits the 418-node component into components of 268 and 149 nodes. These two removals reduce normalized GCC from 0.2635 to 0.1340, producing the sharp descent in Fig.~\ref{fig:ablation}(b). The resulting insight is degree is to use early hub removal to expose structural bottlenecks and then prioritize low-degree separators through cross-boundary reach, bridge evidence. 

\section{Limitations and Ethical Considerations}

AutoSND is validated mainly on network dismantling. Its influence-maximization extension in Appendix~\ref{app:scaling} supports feasibility on one additional task but not broad generality. Future work should test more graph objectives and domains. Real-network use must respect data authorization, privacy, and domain constraints.

\section{Generative AI Usage}

Generative AI is an explicit component of AutoSND, generating candidate programs and inducing structural policies (Section~\ref{sec:method}). It also assisted with code drafting and debugging, figure generation, English translation, and language polishing. 

\section{Conclusion}

AutoSND converts candidate quality, candidate runtime, execution states, and code operations into an explicit structural policy that conditions subsequent program search. AutoSND-Q/S complete all 12 real networks with competitive quality and low candidate runtime and also return valid complete sequences on three large networks. Ablations show that Stage II mainly supports cross-network executability, while structure-diverse Stage III seeds preserve validity and quality. The final programs expose a reusable pattern: maintain residual degree as the backbone, adjust ordering with bounded local signals, and restrict updates to affected neighborhoods. AutoSND therefore connects execution evidence, structural knowledge, and conditional generation for interpretable network dismantling heuristic discovery. Moreover, AutoSND provides a data-driven route for formulating and testing executable structural hypotheses in network science.

\bibliographystyle{ACM-Reference-Format}
\bibliography{sample-base}

\appendix

\section{Search Evaluation, Candidate Selection, and Validity Specification}
\label{app:search}

This appendix provides the search evaluation, deterministic selection rules, and configuration fields required for reproduction. The actual budgets and proxy graph used in Section~\ref{sec:experiments} are given in Section~\ref{sec:search-settings}, and candidate sources are given in Appendix~\ref{app:evidence}.

\subsection{Search Stage Evaluation}
\label{app:search-evaluation}

\subsubsection{Hyperparameters, Initial Values, and Archive Budgets}

\begin{table*}[t]
\caption{Unified configuration for AutoSND search, archiving, and final candidate selection.}
\label{tab:config}
\centering
\scriptsize
\begin{tabular}{llll}
\toprule
Category & Parameter & Value or initial value & Role\\
\midrule
Initial program & $h_0$ & original HDA & Initialize the Stage I search tree with dynamic residual degree\\
Three stage budgets & $(B_1,B_2,B_3)$ & $(300,10,200)$ & I/III count candidates, II counts induction calls\\
Single generation & candidates per call & 1 & One complete candidate per LLM call\\
Parallelism/timeout & workers, $\tau_{\max}$ & 4, 90 s & Parallel LLM requests, for each candidate search limit\\
Stage I weights & $(\alpha,\beta)$ & $(0.80,0.20)$ & ANC@30\% (GCC) and candidate runtime weights\\
Parent expansion & $(\lambda_1,\lambda_3)$ & $(0.10,0.10)$ & Low expansion count reward in Stages I/III\\
Q branch & $(\alpha_Q,\beta_Q),K_Q$ & $(0.90,0.10),8$ & Quality oriented seeds\\
S branch & $(\alpha_S,\beta_S),K_S,r_S$ & $(0.60,0.40),6,0.55$ & Quality gate, then speed oriented selection\\
B branch & $(\alpha_B,\beta_B),K_B$ & $(0.80,0.20),6$ & Quality and candidate runtime balance and structural rotation\\
R branch & $(\alpha_R,\beta_R),K_R,\gamma_R$ & $(0.90,0.10),4,0.75$ & Q seeds plus candidates at/above candidate runtime quantile 0.75\\
Stage III expansion & Q/S/B/R & $(60,60,50,30)$ & Fixed allocation of 200 expansions\\
Stage II summary & TopValid, FeatureRows & 20, 80 & Top 20 valid records and first 80 code features\\
Stage II merge & max proposals & 10 & Merge at most 10 parsable proposals in call order\\
Policy initialization & neighbor cap, two-hop cap & 64, 128 & Initial access caps of \texttt{BasePolicy}\\
Generation model & model, effort, temperature & \texttt{gpt-5.5}, \texttt{none}, 0.2 & Unified AutoSND/AHD generation setting\\
Source search & search seed & 42 & Source search of frozen Q/S candidates\\
\bottomrule
\end{tabular}
\end{table*}

Table~\ref{tab:config} gives the unified search configuration. The run archive covers candidate records from all three stages, Stage II prompts and raw responses, and complete parent to child relationships. The top 20 and first 80 records form a bounded evidence summary for one Stage II prompt. Final candidate identities, source nodes, and the proxy graph are given in Appendix~\ref{app:evidence}.

\subsubsection{Candidate Evaluation and Normalization}
\label{app:candidate-normalization}

The search evaluator first performs deterministic normalization of candidate output: it retains the first occurrence of each node belonging to $V(G)$, in occurrence order, to obtain $\widehat\pi_h$. This relaxed normalization is used only during search. Deployment uses the strict full permutation check in Appendix~\ref{app:validity}.

For proxy graph $G$, let $N=|V(G)|$ and $K=\operatorname{round}(0.30N)$, and let $s_k$ be the GCC size of the residual graph after removing the first $k$ nodes of $\widehat\pi_h$. The accumulated normalized connectivity over the first 30\% is
\begin{equation}
C_G(h)=\operatorname{ANC}@30\%(\mathrm{GCC},h)
=\frac{1}{K}\sum_{k=1}^{K}\frac{s_k}{\sigma_{\mathrm{GCC}}(G)}.
\end{equation}
All proxy graphs used in the search are connected, so $\sigma_{\mathrm{GCC}}(G)=N$, consistent with the node count normalization in the final candidate implementation. Let the candidate runtime for generating a removal sequence on $G$ be
\begin{equation}
\tau_G(h)=t_{\mathrm{end}}(h,G)-t_{\mathrm{start}}(h,G).
\end{equation}
On the proxy graph set $\mathcal G_p$, both search stage quality and candidate runtime use macro averages:
\begin{equation}
a(h)=\frac{1}{|\mathcal G_p|}\sum_{G\in\mathcal G_p}C_G(h),\quad
\tau(h)=\frac{1}{|\mathcal G_p|}\sum_{G\in\mathcal G_p}\tau_G(h).
\end{equation}
If any proxy graph produces a syntax, interface, runtime, or timeout error, the candidate is invalid. Only valid candidates enter the quality and candidate runtime comparison. For $M$ valid candidates in the current stage, both metrics are minimized, and the normalized rank is
\begin{equation}
\rho(x_i)=1-\frac{\operatorname{rank}_{\rm asc}(x_i)-1}{M-1}.
\end{equation}
When $M=1$, $\rho(x_i)=1$. Given quality and candidate runtime weights $\alpha+\beta=1$, the unified candidate score is
\begin{equation}
\mathbf f_{\alpha,\beta}(h_i)
=\alpha\rho\!\left(a(h_i)\right)
+\beta\rho\!\left(\tau(h_i)\right).
\end{equation}
Stages I and III recompute ranks within their respective valid candidate sets. Parent candidate priority adds an expansion term to this score, and the quality and candidate runtime preferences of the four Stage III seed types are given below. This evaluation is used only for search selection. Final results are still reevaluated using complete sequence ANC (GCC).

\subsection{Complete Stage III Seed Selection Procedure}
\label{app:seed-selection}

Stage III selects seeds directly from valid Stage I candidates. $\Pi$ conditions subsequent generation but does not change seed scores. Let $\rho_1(a_i),\rho_1(\tau_i)\in[0,1]$ be the normalized ranks recomputed on $\mathcal V_1$, with 1 indicating the best value. Q selects the quality oriented top candidates:
\begin{equation}
\mathcal H_{\mathrm Q}
=\operatorname{Top}_{K_{\mathrm Q}}
\left(\mathcal V_1,\alpha_{\mathrm Q}\rho_1(a)+\beta_{\mathrm Q}\rho_1(\tau)\right).
\end{equation}
S first applies the fixed normalized quality-rank threshold $r_{\mathrm S}=0.55$, recomputes ranks $\rho_{\mathrm S}$ within the retained pool, and then prioritizes candidate runtime:
\begin{align}
\mathcal V_{\mathrm S}
&=\{h_i\in\mathcal V_1:\rho_1(a_i)\geq r_{\mathrm S}\},\\
\mathcal H_{\mathrm S}
&=\operatorname{Top}_{K_{\mathrm S}}
\left(\mathcal V_{\mathrm S},
\alpha_{\mathrm S}\rho_{\mathrm S}(a)+\beta_{\mathrm S}\rho_{\mathrm S}(\tau)\right).
\end{align}
B ranks the complete valid pool with balanced weights and rotates across the structural signature groups $g(z_i)$:
\begin{equation}
\mathcal H_{\mathrm B}
=\operatorname{DiverseTop}_{K_{\mathrm B}}
\left(\mathcal V_1,
\alpha_{\mathrm B}\rho_1(a)+\beta_{\mathrm B}\rho_1(\tau),g(z_i)\right).
\end{equation}
Finally, the R pool combines Q seeds with candidates at or above the candidate runtime quantile $\gamma_{\mathrm R}$ and reuses the Q score:
\begin{align}
\mathcal V_{\mathrm R}
&=\mathcal H_{\mathrm Q}\cup
\left\{h_i\in\mathcal V_1:
\tau_i\geq
\operatorname{Quantile}_{\gamma_{\mathrm R}}
\left(\{\tau_j:h_j\in\mathcal V_1\}\right)\right\},\\
\mathcal H_{\mathrm R}
&=\operatorname{Top}_{K_{\mathrm R}}
\left(\mathcal V_{\mathrm R},
\alpha_{\mathrm Q}\rho_1(a)+\beta_{\mathrm Q}\rho_1(\tau)\right).
\end{align}
The four sets initialize separate Q, S, B, and R branches. A candidate selected by multiple rules retains each branch assignment. Unique seed counts are reported only as a diagnostic.

\begin{table}[t]
\caption{Candidate pool, ranking rule, and seed quota of the four Stage III branches.}
\label{tab:seeds}
\centering
\scriptsize
\begin{tabular}{p{0.15\columnwidth}p{0.65\columnwidth}r}
\toprule
Branch & Candidate pool and ranking & Quota\\
\midrule
Q & Quality oriented $\mathbf f_Q$ & 8\\
S & Retain $\rho_1(a)\geq r_{\mathrm S}$ with $r_{\mathrm S}=0.55$, then speed oriented $\mathbf f_S$ & 6\\
B & Balanced $\mathbf f_B$, then rotate by $g(z)$ & 6\\
R & Union of Q seeds and $\tau\geq Q_{0.75}(\tau)$, then $\mathbf f_Q$ & 4\\
\bottomrule
\end{tabular}
\end{table}

\begin{table}[t]
\caption{Stage I candidate pool and seed deduplication statistics of the search seed 42 source search.}
\label{tab:seedstats}
\centering
\scriptsize
\begin{tabular}{lrrrrr}
\toprule
Search & Valid & S pool & R slow & R union & Unique\\
\midrule
AutoSND-Q/S & 299 & 135 & 75 & 83 & 11\\
\bottomrule
\end{tabular}
\end{table}

The same candidate may initialize multiple branches, for which branch prompts, parent to child relationships, and budgets are separately recorded. Branch expansion quotas are given in Table~\ref{tab:config}. In each round, a parent is first selected from valid tree nodes in the same branch. When this pool is empty, the global valid parent pool is used.

\subsection{ANC (GCC) and Candidate Runtime Frontier and Final Candidate Determination}
\label{app:candidate-selection}

Let $\mathcal V_3$ be the valid candidate set of Stage III. If $h_i,h_j\in\mathcal V_3$ satisfy $a(h_i)\leq a(h_j)$ and $\tau(h_i)\leq\tau(h_j)$, with at least one strict inequality, then $h_i$ dominates $h_j$. Nondominated candidates form $\mathcal F_3$. The quality point within a run is
\begin{equation}
h_Q^{\rm run}=\arg\min_{h\in\mathcal F_3}(a(h),\tau(h)).
\end{equation}
For the speed point, the normalized quality rank $\rho_{\mathcal F_3}(a(h))$ is recomputed on $\mathcal F_3$. Candidates satisfying $\rho_{\mathcal F_3}(a(h))\geq r_{\mathrm S}$, where $r_{\mathrm S}=0.55$, are retained and then selected in ascending lexicographic order of $(\tau(h),a(h))$. When these values are exactly equal, archive order breaks the tie. Both representative candidates reported in this paper come from the AutoSND Stage III source archive on one \texttt{Powerlaw\_500} proxy graph with search seed 42 and are frozen according to the ANC@30\% (GCC)--candidate runtime frontier:
\begin{equation}
\mathcal H_{\rm final}
=\{\texttt{cd9e3818033d},\texttt{0ade8d3405c2}\}.
\end{equation}
\texttt{cd9e3818033d} is the quality side representative AutoSND-Q, and \texttt{0ade8d3405c2} is the speed side representative AutoSND-S. Q/S are role labels determined on the search proxy graph at freezing time. Results on the 12 main graphs and large real networks after freezing are only external reevaluations and do not change candidate identity or role. Source nodes and archive records are given in Appendix~\ref{app:evidence}.

\subsection{Validity, Coverage, and Diagnostic Quantities}
\label{app:validity}

\begin{itemize}
\item A search candidate is valid if it passes syntax, interface, runtime, and timeout checks on all proxy graphs. Its output is normalized as in Appendix~\ref{app:candidate-normalization} and truncated to the first $K_G$ steps.
\item A main text evaluation input is complete if it returns within 600 s a sequence of length $N$, without duplicates, whose node set equals $V$.
\item A million node stress test succeeds if the synthetic graph run returns $N$ distinct node values. This test does not use the 600 s hard limit for comparisons in the main text for each input. The three public large graphs additionally undergo strict node set matching, complete trajectory checking, and independent ANC (GCC) verification.
\end{itemize}

Candidate validity and complete sequence coverage are computed as
\begin{equation}
\operatorname{ValidRate}=\frac{N_{\mathrm{valid}}}{B}\times100\%,
\quad
\operatorname{Coverage}=\frac{N_{\mathrm{complete}}}{|\mathcal G_{\mathrm{test}}|}\times100\%.
\end{equation}
For AutoSND, $N_{\mathrm{valid}}=|\mathcal V_1|+|\mathcal V_3|$ and $B=B_1+B_3$.

\section{Structural Policy Compilation and Prompts}
\label{app:implementation}

This appendix provides the three stage data flow consistent with the main text and further explains how Stage II aligns candidate code structures with execution results, forms a structural policy, and writes it into the Stage III candidate generation condition. Raw prompts, responses, the compiled structural policy, and the complete text entering Stage III are saved with each run.

\subsection{Three Stage Pseudocode}

\begin{verbatim}
Input: h0, proxy graphs Gp, budgets B1/B2/B3,
    evaluator, and candidate interface
Stage I: Broad Tree Search
 evaluate h0 and initialize the search tree
 for t = 1 ... B1:
  rank valid nodes by the fixed search score
  select a parent with exploration bonus
  request one complete degree_order(G)
  run the code guard and proxy evaluation
  archive code, parent, prompt, status, metrics,
   and candidate runtime
 archive Stage I records E1
Stage II: Execution-to-Evidence Structural
     Policy Induction
 for each Stage I record ei in E1:
  zi <- LLM-Struct(hi), merge zi into ei
 C1 <- positive and negative contrasts from E1
 T1 <- structural statistics aligned with quality,
    candidate runtime, and status
 for j = 1 ... B2:
  qj <- LLM(C1,T1), archive prompt/proposal
  append valid qj to Q
 Pi <- Compile(Q,C1,T1), archive Pi
Stage III: Policy-Conditioned Tree Search
 select Q/S/B/R seeds from valid Stage I records
 for t = 1 ... B3:
  select a branch and parent under fixed schedule
  prompt <- role + parent code/metrics + complete Pi
  request one complete degree_order(G)
  reject static high cost pattern violations
  evaluate/archive child, valid children may expand
 construct the quality and candidate runtime frontier
 freeze Q/S representatives on Powerlaw_500
\end{verbatim}

The structural signature
\begin{equation}
g(z_i)=\operatorname{Join}\!\left(\{x\in\mathcal K:z_i[x]=1\}\right)
\end{equation}
is used only for diversified seed selection in the B branch, where
\begin{equation}
\begin{aligned}
\mathcal K=\{&
\texttt{degree},\texttt{frontier},\texttt{weak\_tie},\\
&\texttt{two\_hop},\texttt{boundary},\texttt{redundancy},
\texttt{phase},\texttt{heap}\}.
\end{aligned}
\end{equation}
An empty signature is denoted by \texttt{plain}.

\subsection{Code Features and Execution Summary}

Stage II analyzes Stage I candidates in a unified structural description space $\Omega$. For each candidate $h_i$, structural description $z_i$ summarizes its code structure family, local structural signals, neighborhood access, state maintenance, update scope, and high cost operations and is merged with the corresponding execution record. To avoid conflict with the main text candidate symbol $h_i$, $z_i$ uses named fields:
\begin{equation}
z_i:\mathcal K_{\mathrm{full}}\rightarrow\{0,1\}\cup\mathbb N_0,
\end{equation}
where
\begin{equation}
\begin{aligned}
\mathcal K_{\mathrm{full}}=
\{&\texttt{degree},\texttt{frontier},\texttt{weak\_tie},\\
&\texttt{two\_hop},\texttt{boundary},\texttt{redundancy},
\texttt{phase},\texttt{heap},\\
&\texttt{component\_refresh},\texttt{global\_recompute},\\
&\texttt{unbounded\_two\_hop},\texttt{neighbors\_calls}\}.
\end{aligned}
\end{equation}
These fields denote structures such as the residual degree backbone, frontier, weak connection, two-hop access, boundary, redundancy, stage scheduling, heap update, at each step component refresh, global recomputation, unbounded two-hop access, and direct neighbor access. Structural descriptions are uniformly mapped to code structure families such as \texttt{component refresh}, \texttt{two-hop boundary}, \texttt{redundancy aware}, \texttt{local heap update}, and \texttt{degree local}, supporting comparison across candidates and structurally diversified seed selection in the B branch.

The augmented execution archive first forms positive and negative contrastive evidence $\mathcal C_1$. Positive examples come from candidates leading in quality and candidate runtime. Negative examples are divided into poor quality, slow, and execution failure classes according to their primary disadvantage. Stage II then calculates the occurrence proportions of each structural item and sufficiently sampled combination in positive and negative candidates, and summarizes the corresponding ANC@30\% (GCC), candidate runtime, execution status, and parsable neighborhood access and state update scopes, forming historical statistics $\mathcal T_1$.

The execution summary entering the Stage II prompt consists of contrastive candidates and their execution results, historical statistics of structural families and items, and summaries of access and update scopes. It aligns candidate execution results with code structures, allowing the preferred structures, avoided structures, and scope recommendations proposed by the LLM to be traced back to the same Stage I archive.

\subsection{Proposal Interface and Deterministic Compilation}

Each Stage II call is required to return a JSON object with the fields \texttt{allowed\_signals}, \texttt{preferred\_families}, \texttt{pruned\_families}, and \texttt{cap\_bounds}. The remaining required fields are \texttt{update\_bounds}, \texttt{forbidden\_patterns}, and \texttt{stage3\_prompt\_contract}. Allowed signals, preferred structural families, caps, supported update fields, and forbidden patterns participate in evidence adaptive merging. Pruned families and the overall Stage III generation constraints are fixed by a deterministic compiler so that runs share the same safety boundary. If a response cannot be parsed as JSON, its raw text and error are archived, and it does not enter the merge.

\section{Data Sources, Complete Sequence Coverage, and For Each Graph AutoSND Results}
\label{app:data}

\subsection{Dataset Sizes}
\label{app:dataset-sizes}

\begin{table*}[t]
\caption{Sizes, primary categories, and experimental uses of real networks.}
\label{tab:datasets}
\centering
\scriptsize
\begin{tabular}{lrrll}
\toprule
Dataset & $N$ & $M$ & Primary category & Use\\
\midrule
CEnew & 453 & 2,025 & Main test & Main comparison, code component ablation\\
crime & 829 & 1,473 & Main test & Main comparison, code component ablation\\
email & 1,133 & 5,451 & Main test & Main comparison, code component ablation\\
PH & 2,000 & 16,098 & Archive/ablation & Full archive, original ablation input\\
hamster & 2,000 & 16,098 & Main test & Main comparison, Fig.~\ref{fig:ablation} curve\\
Yeast & 2,375 & 11,693 & Main test & Main comparison, code component ablation\\
Collaboration & 4,158 & 13,422 & Main test & Main comparison, code component ablation\\
GrQC & 4,158 & 13,422 & Archive/ablation & Full archive, code component ablation\\
HI-II-14 & 4,165 & 13,087 & Main test & Main comparison, ablation reevaluation\\
Grid & 4,941 & 6,594 & Main test & Main comparison, code component ablation\\
HepPh & 11,204 & 117,619 & Main test & Main comparison, code component ablation\\
condmat & 21,363 & 91,286 & Main test & Main comparison, code component ablation\\
Digg & 29,652 & 84,781 & Main test & Main comparison, ablation reevaluation\\
Enron & 33,696 & 180,811 & Archive & Full result archive\\
Gnutella31 & 62,561 & 147,878 & Main test & Main comparison, ablation and accelerated reevaluation\\
Epinions & 75,877 & 405,739 & Archive & Full result archive\\
Facebook & 63,392 & 816,831 & Large validation & Fig.~\ref{fig:large} and complete results\\
YouTube & 1,134,890 & 2,987,624 & Large validation & Fig.~\ref{fig:large} and complete results\\
Flickr & 1,624,991 & 15,473,043 & Large validation & Fig.~\ref{fig:large} and complete results\\
\bottomrule
\end{tabular}
\end{table*}

Edge by edge provenance shows that CEnew matches the accompanying data of the published method~\cite{guo2020}. Email corresponds to a public email network~\cite{guimera2003,kunegis2013}. Hamster corresponds to the largest connected component of the original KONECT graph~\cite{kunegis2013}. Condmat, HepPh, and GrQC correspond to the largest connected components of SNAP \texttt{ca-CondMat}, \texttt{ca-HepPh}, and \texttt{ca-GrQc} after removing self loops~\cite{leskovec2014}. Grid matches SuiteSparse \texttt{Newman/power}~\cite{davis2011}. Yeast matches the yeast PPI snapshot used by WalkPooling~\cite{pan2021,vonmering2002}. HI-II-14, Digg, Enron, Gnutella31, and Epinions use locally archived undirected simple graph versions, whose file hashes and actual node/edge counts are recorded in the run manifest. Gnutella31 is reported according to the local file count rather than replacing it with the size shown on a public page with the same name.

PH and hamster have identical undirected simple edge sets, and Collaboration and GrQC are isomorphic. \texttt{crime} is obtained by folding the identifier space of KONECT bipartite data~\cite{kunegis2013}. The 12 main graphs are CEnew, Collaboration, condmat, crime, email, Grid, hamster, HepPh, Yeast, HI-II-14, Digg, and Gnutella31. Real graphs not included in the main mean are still fully recorded in Appendix~\ref{app:fullresults}.

\paragraph{Boundary between proxy/ablation graphs and main test graphs.}
The synthetic graph \texttt{Powerlaw\_500} contains 500 nodes and 943 edges and is used as the search proxy and to freeze Q/S roles. It is not a real network in Table~\ref{tab:datasets} and is not included in the mean over the 12 main graphs. Results for each graph for PH, GrQC, and \texttt{Powerlaw\_500} are retained only in the complete records of Appendix~\ref{app:fullresults} and do not form a separate 12 graph statistic. The 12 main graphs and three large real networks are used for unified reevaluation only after candidate identities and Q/S roles have been frozen.

\subsection{Baseline Reevaluation Rules}

All methods follow the complete sequence evaluation, completion criterion, and candidate runtime protocol in Section~\ref{sec:evaluation-metrics}. Native baselines use Python implementations. AHD baselines are mechanism level local reproductions under the unified network dismantling interface.

The AI agent controls use Codex 26.721.41059 and the same proxy graph, \texttt{degree\_order(G)} interface, evaluator, and $300+200=500$ candidate budget as AutoSND. Codex-Direct uses independent subagents with reasoning effort \texttt{low} and shares no candidate history. Codex-Evo first asks Codex with reasoning effort \texttt{xhigh} to design an evolutionary scheme using the prompt ``Please design an evolutionary algorithm scheme for solving the network dismantling problem''. \texttt{low} effort subagents then independently generate candidates under this scheme.

\begin{table*}[t]
\caption{Local reproduction scope, main differences, and complete sequence coverage of AHD baselines.}
\label{tab:reproduction}
\centering
\scriptsize
\resizebox{\textwidth}{!}{%
\begin{tabular}{llllr}
\toprule
Method & Basis & Core mechanism implemented & Implementation under unified interface & Coverage\\
\midrule
FunSearch & Paper/public structure~\cite{romeraparedes2024} & Four islands, elite parents, independent mutation & Local four island controller with unified prompt/evaluator & 100\%\\
Clade-AHD & Paper/public structure~\cite{chen2026} & Structural signature clades, Thompson sampling, elite selection & Local clade metadata/selector with unified evaluator & 100\%\\
ERA & Paper/public structure~\cite{zheng2025era} & PUCT parent selection from scores/visits & Unified task adapter, budget, and evaluator & 100\%\\
MCTS-AHD & Paper/public structure~\cite{zheng2025} & UCT, initialization, structural/parameter mutation, crossover & One complete program per candidate, same limit & 100\%\\
AlphaEvolve & Paper/OpenEvolve reference~\cite{novikov2025} & Code feature bins, archive/niche, quality and novelty balance & Local archive/niche controller from description/reference & 100\%\\
LLM4CN & Paper/author repository~\cite{zhang2025llm4cn} & Population classification, capacity, sampling, connectivity fitness & Clean room, \texttt{degree\_order(G)}, unified proxy/budget & 75.0\%\\
EoH & Paper/public structure~\cite{liu2024eoh} & Idea and code coevolution, crossover, two mutations & Operators retained, unified prompt/evaluator & 100\%\\
ReEvo & Paper/public structure~\cite{ye2024} & Short term and long term reflection guided mutation/crossover & Reflections from local logs, unified interface/evaluator & 91.7\%\\
HiFo-Prompt & Paper/public structure~\cite{mao2024} & Hindsight insight pool and foresight navigator & Insight pool from local logs, unified task/budget & 91.7\%\\
HSEvo & Paper/public structure~\cite{liu2024} & Harmony memory, pitch adjustment, diversity crossover & Search pressure retained, local archive/unified evaluator & 75.0\%\\
\bottomrule
\end{tabular}}
\end{table*}

Each AHD controller saves the parent candidate, generation operator, prompt, raw response, candidate source, validity, proxy metrics, and candidate runtime. Invalid responses also consume one budget unit. Candidate validity in Table~\ref{tab:maincomparison} is calculated from complete search logs, and final candidate ANC (GCC) and candidate runtime come from unified replay on the 12 main graphs. AlphaEvolve in the table denotes the local archive/niche controller listed in Table~\ref{tab:reproduction}.

In reevaluation on the 12 main graphs, FunSearch, Clade-AHD, ERA, MCTS-AHD, AlphaEvolve, and EoH complete 12/12. ReEvo and HiFo-Prompt complete 11/12. LLM4CN and HSEvo complete 9/12. Codex-Evo and Codex-Direct complete 8/12 and 3/12, respectively. Local runs of CoreHD, DC, CI, KCore, and CLUC complete 12/12. Rule preserving accelerated reevaluations of HDA, GND, MinSum, and BPD complete all 48 method and input cells. FINDER has quality coverage of 12/12 and candidate runtime coverage of 11/12. Local reevaluations of AutoSND-Q/S both complete 12/12.

HDA, MinSum, and BPD pass element by element complete sequence identity checks on CEnew and HI-II-14 and ANC (GCC) verification on the 12 main graphs. GND is reevaluated using a fixed nonconstant initial vector for \texttt{eigsh} and passes two complete sequence identity checks. FINDER candidate runtimes for Digg and Gnutella31 come from \texttt{uniform\_cost/solution\_time.csv}. Its mean candidate runtime is calculated over the 11 main test graphs with same source records.

DC sorts by static degree in descending order. local KCore sorts by static core number, degree, and node identifier. local CLUC uses $10^{-C_v}\sum_{u\in N(v)}(d_u+1)$, where $C_v$ is the local clustering coefficient of node $v$. A method that does not complete all inputs is not assigned a mean.

\begin{table}[t]
\caption{Complete sequence coverage of different candidate generation methods.}
\label{tab:generationcoverage}
\centering
\scriptsize
\begin{tabular}{llr}
\toprule
Method & Candidate generation & Coverage\\
\midrule
Codex-Direct& 500 independent direct candidates & 25.0\%\\
Codex-Evo & 500 simplified evolution candidates & 66.7\%\\
\textbf{AutoSND-Q} & Three stage AutoSND search & \textbf{100\%}\\
\textbf{AutoSND-S} & Three stage AutoSND search & \textbf{100\%}\\
\bottomrule
\end{tabular}
\end{table}
This table only compares complete sequence coverage under the current fixed generation and selection implementations. It does not enter quality ranking or a single factor judgment.

\subsection{For Each Graph Results of AutoSND-Q/S}

\begin{table}[t]
\caption{Supplementary for each graph results of frozen AutoSND-Q/S on small and medium inputs.}
\label{tab:pergraph}
\centering
\scriptsize
\begin{tabular}{lrrrr}
\toprule
Dataset & Q ANC (GCC) & \shortstack{Q candidate\\runtime} & S ANC (GCC) & \shortstack{S candidate\\runtime}\\
\midrule
CEnew & 0.0966 & 0.033 & 0.0980 & 0.034\\
condmat & 0.1077 & 3.868 & 0.1083 & 3.841\\
email & 0.2167 & 0.181 & 0.2170 & 0.180\\
Grid & 0.0387 & 0.117 & 0.0458 & 0.163\\
GrQC & 0.0786 & 0.386 & 0.0755 & 0.388\\
hamster & 0.1519 & 0.823 & 0.1499 & 0.599\\
HepPh & 0.1411 & 23.464 & 0.1367 & 6.636\\
Powerlaw\_500 & 0.0928 & 0.017 & 0.0997 & 0.018\\
Yeast & 0.1210 & 0.531 & 0.1142 & 0.392\\
\textbf{Mean} & \textbf{0.1161} & 3.269 & \textbf{0.1161} & 1.361\\
\bottomrule
\end{tabular}
\end{table}

This table supplements the results of the two frozen candidates on small and medium inputs and does not participate in Q/S role determination. Q/S roles are frozen only according to the \texttt{Powerlaw\_500} search archive. Collaboration and crime results are in Table~\ref{tab:maincomparison}. PH and other for each graph results not entering the main mean are retained in Appendix~\ref{app:fullresults}.

\subsection{Figures and Source Tables}

All figures and tables in the main text are redrawn from UTF 8 source tables. Search stage ANC@30\% (GCC) and Fig.~\ref{fig:trajectory} cover only the removal ratio interval $[0,0.30]$, whereas complete sequence conclusions use for each input replay results. Table~\ref{tab:maincomparison}, the left panel of Fig.~\ref{fig:ablation}, and Appendix~\ref{app:component-ablation} all use the 12 main graphs as their data boundary. Candidate runtimes come from the corresponding execution records. Major source tables, run identifiers, and file checksums are jointly indexed by the accompanying reproduction manifest.

\section{Structures, Ablations, and Scaling Results of AutoSND-Q/S}
\label{app:scaling}

\subsection{Key Code Component Variants}
\label{app:component-ablation}

Each variant replays a complete removal sequence on the 12 main graphs. $\Delta=\operatorname{ANC}(\mathrm{GCC})_{\mathrm{variant}}-\operatorname{ANC}(\mathrm{GCC})_{\mathrm{full}}$. A positive value means that the complete version is better. Missing cells for HI-II-14, Digg, and Gnutella31 are concurrently reevaluated under the 600 s limit using the same component variants, and all 42 cells return valid complete sequences.

\begin{table}[t]
\caption{Complete sequence ablation of main AutoSND-Q/S code components.}
\centering
\scriptsize
\begin{tabular}{llrr}
\toprule
Candidate & Variant & Mean ANC (GCC) & $\Delta$ ANC (GCC)\\
\midrule
Q & Full & \textbf{0.1099} & 0\\
Q & Residual degree only & 0.1254 & +0.0155\\
Q & Remove residual degree main term & 0.1259 & +0.0160\\
Q & Remove redundancy penalty & 0.1122 & +0.0023\\
Q & Remove weak connection and bridge & 0.1116 & +0.0017\\
Q & Remove boundary & 0.1104 & +0.0005\\
Q & Remove bounded two-hop & 0.1101 & +0.0003\\
S & Full & \textbf{0.1093} & 0\\
S & Residual degree only & 0.1254 & +0.0161\\
S & Remove residual degree main term & 0.1111 & +0.0019\\
S & Remove redundancy penalty & 0.1112 & +0.0019\\
S & Remove boundary ratio & 0.1106 & +0.0013\\
S & Remove weak connection and bridge & 0.1100 & +0.0007\\
S & Remove frontier and bounded two-hop & 0.1101 & +0.0009\\
\bottomrule
\end{tabular}
\end{table}

Relative to using only residual degree, complete Q/S reduce mean ANC (GCC) by 12.4\% and 12.9\%, respectively. Major local terms including redundancy, weak connection/bridge, boundary, and bounded two-hop access also yield consistent ANC (GCC) improvements.

\subsection{Synthetic Graph Scaling and Million Node Stress Tests}

Scaling tests cover power law, Erdős--Rényi (ER), Watts--Strogatz (WS), and stochastic block model (SBM) graphs from 500 to 10,000 nodes with synthetic graph seeds 42--44. Across five methods, four sizes, four graph families, and three seeds, all 240 cells complete ANC (GCC) calculation. At 10,000 nodes, the four family, three seed mean ANC (GCC) values of AutoSND-Q and AutoSND-S are 0.240982 and 0.249783, respectively, both lower than the three residual degree baselines. The corresponding mean candidate runtimes are 2.888 s and 1.660 s. Both methods subsequently return sequences containing $N$ distinct node values with a 100\% completion rate on million node synthetic Powerlaw, ER, WS, and SBM graphs.

\begin{table}[t]
\caption{Complete evaluation with 10,000 nodes and million node stress tests.}
\centering
\scriptsize
\begin{tabular}{lrrr}
\toprule
Method & 10k mean ANC (GCC) & 10k runtime (s) & Million node completion\\
\midrule
\textbf{AutoSND-Q} & \textbf{0.240982} & 2.888 & \textbf{100\%}\\
\textbf{AutoSND-S} & 0.249783 & 1.660 & \textbf{100\%}\\
HDA & 0.284135 & 5.785 & --\\
fast HDA & 0.284867 & 0.192 & --\\
fast CoreHD & 0.283920 & 0.166 & --\\
\bottomrule
\end{tabular}
\end{table}
Q obtains the lowest ANC (GCC) on the 10k graphs, while S obtains the second best ANC (GCC) with a lower mean candidate runtime. Both successfully return full length node sequences in all runs over the four million node graph families.

\subsection{Candidate Runtime by Million Node Graph Family}

For complete evaluation at 500--10,000 nodes, Powerlaw samples degrees according to $p(k)\propto k^{-2.5}$ for $k\in[2,\lfloor4\sqrt N\rfloor]$ and then constructs a simplified configuration model~\cite{molloy1995}. Million node stress runs use the cheaper to generate BA$(N,2)$ model~\cite{barabasi1999}. ER uses $p=\min(0.05,6/(N-1))$~\cite{erdos1959}. WS uses even $k=\min(8,\max(2,2\lfloor N/50\rfloor))$ and rewiring rate 0.08~\cite{watts1998}. SBM uses four nearly equal blocks, $p_{\rm in}=\min(0.08,10/(N/4))$, and $p_{\rm out}=0.05p_{\rm in}$~\cite{holland1983}. For $N\geq10^5$, edges are sampled according to a target edge count to control generation overhead. All graphs have self loops removed, are converted to undirected graphs, and have components connected by single edges.

\begin{table}[t]
\caption{Candidate runtime and completion rate by graph family in million node stress runs.}
\centering
\scriptsize
\resizebox{\columnwidth}{!}{%
\begin{tabular}{lrrrrrr}
\toprule
Family & Q mean & Q max & S mean & S max & Q comp. & S comp.\\
\midrule
Powerlaw & 158.022 & 161.098 & 148.581 & 153.318 & 100\% & 100\%\\
ER & 437.049 & 447.515 & 371.543 & 382.876 & 100\% & 100\%\\
WS & 202.324 & 204.742 & 197.264 & 200.290 & 100\% & 100\%\\
SBM & 1774.275 & 1781.799 & 931.536 & 933.166 & 100\% & 100\%\\
\textbf{All} & \textbf{642.917} & \textbf{1781.799} & \textbf{412.231} & \textbf{933.166} & \textbf{100\%} & \textbf{100\%}\\
\bottomrule
\end{tabular}}
\end{table}
Each stress test cell saves an output containing $N$ distinct node values. The next subsection additionally performs strict node set matching, complete trajectory checking, and independent ANC (GCC) verification on three public large graphs.

\subsection{Public Large Graph Evaluation and Reproduction Settings}
\label{app:large-reproduction}

External graph runs use the two final candidates. Their identifiers are Q (\texttt{cd9e3818033d}) and S (\texttt{0ade8d3405c2}). All six runs return strict node permutations and save stepwise GCC curves and hashes of candidate and graph files. The maximum absolute ANC (GCC) difference under independent reverse union find recomputation is 0. Every curve contains $N$ rows, is monotonically nonincreasing, and ends at 0. The environment is Windows 11, Python 3.12.9, and NetworkX 3.4.2 on a machine with an Intel Core Ultra 9 275HX and 64 GB memory.

The FINDER comparison uses only the without reinsertion curves corresponding to Figure 5g of its paper: \nolinkurl{MaxCCList_Strategy_facebook-links.txt}, \nolinkurl{MaxCCList_Strategy_com-youtube.txt}, and \nolinkurl{MaxCCList_Strategy_flickr-links.txt}~\cite{fan2020}. Each released curve contains $N$ normalized GCC values. Multiplying the file mean by 100 gives 26.843389, 2.366464, and 5.455436, reproducing the paper's 26.84, 2.37, and 5.46. \nolinkurl{MaxCCList_Strategy_reinsert1_*} belongs to another variant and is not included in the main text quality figure.

The released curves are recorded from an initial value of 1.0 to $1/N$ when one node remains, whereas the unified evaluator in Section~\ref{sec:evaluation-metrics} records from the first removal to a terminal value of 0. Shifting the FINDER curves to the latter convention is equivalent to subtracting $100/N$, yielding 26.841812, 2.366376, and 5.455375, respectively. This does not change the two decimal values in the paper, the three decimal macro mean over the three graphs, or the ranking in Fig.~\ref{fig:large}. The main text therefore uses the values reported in the paper, while this appendix retains the one step storage offset.

\subsection{CoreHD Root Initialization Sensitivity}

To test whether the search depends on HDA initialization, we rerun AutoSND with \texttt{original CoreHD} as the root and summarize search seeds 42 and 43, both of which complete the full search and reevaluation on 12 inputs.
\begin{table}[t]
\caption{Results of CoreHD and CoreHD root AutoSND on 12 inputs.}
\centering
\scriptsize
\begin{tabular}{lrrrr}
\toprule
Method & Stage I valid & Stage III valid & Mean ANC (GCC) & vs. CoreHD\\
\midrule
CoreHD & -- & -- & 0.136730 & --\\
CoreHD root AutoSND-Q & 97.3\% & 98.0\% & \textbf{0.135106} & \textbf{$-1.2\%$}\\
\bottomrule
\end{tabular}
\end{table}
CoreHD root AutoSND-Q reduces mean ANC (GCC) by 1.2\%, showing that quality oriented search can continue extracting improvements from different initial heuristics.

\subsection{Cross Task Applicability: Influence Maximization}

This experiment reuses the process of ``Broad Tree Search, Execution-to-Evidence Structural Policy Induction, and Policy-Conditioned Tree Search'' and the three stage budgets in Section~\ref{sec:search-settings}. The candidate interface is changed to \texttt{seed\_order(G,k)}, with fixed $k=12$. The search is initialized with DegreeDiscount IC, and candidates are evaluated using LT quality and candidate runtime. It uses \texttt{Powerlaw\_500} as the proxy graph. Candidates do not internally perform LT/IC Monte Carlo simulation, and results on 12 inputs are used for external evaluation of frozen candidates. Candidate validity is 91.0\% in Stage I and 100.0\% in Stage III, showing that after a task specific structural policy is formed, subsequent search can stably generate executable candidates for this task.

\begin{table}[t]
\caption{LT quality and candidate runtime of fixed influence maximization programs on 12 inputs.}
\centering
\scriptsize
\begin{tabular}{llrr}
\toprule
Group & Method & Mean LT quality & Mean candidate runtime (s)\\
\midrule
Three stage transfer & Transferred-IM-Q & \textbf{0.250583} & \underline{0.0665}\\
Three stage transfer & Transferred-IM-S & \underline{0.250389} & \textbf{0.0628}\\
AHD & MCTS-AHD & \underline{0.250389} & 0.1067\\
AHD & HSEvo & 0.249386 & 0.3195\\
AHD & AlphaEvolve & 0.248897 & 0.1976\\
AHD & FunSearch & 0.248247 & 0.2103\\
AHD & Clade-AHD & 0.247483 & 0.1275\\
Native heuristic & DegreeDiscount IC & 0.243447 & 1.6955\\
\bottomrule
\end{tabular}
\end{table}
Transferred-IM-Q improves mean LT quality over DegreeDiscount IC by 0.007135 (approximately 2.93\%) and over the strongest AHD control by 0.000194 (approximately 0.077\%). Transferred-IM-S and MCTS-AHD have the same quality to six decimal places, while the former reduces mean candidate runtime by approximately 41\%. This experiment validates the transferability of the three stage process and the Stage II evidence and policy interface to influence maximization and extends the current cross task evidence to a second class of complex network optimization problems.

\section{Candidate Sources and Stage II Evidence Tracing}
\label{app:evidence}

\subsection{Candidate Identities and Sources}

\begin{table}[t]
\caption{Identities and sources of the final candidates.}
\centering
\scriptsize
\begin{tabular}{llll}
\toprule
Role & Candidate ID & Archive/node & Freeze proxy\\
\midrule
AutoSND-Q & \texttt{cd9e3818033d} & search seed 42, \texttt{stage3-0044} & \texttt{Powerlaw\_500}\\
AutoSND-S & \texttt{0ade8d3405c2} & search seed 42, \texttt{stage3-0040} & \texttt{Powerlaw\_500}\\
\bottomrule
\end{tabular}
\end{table}
The source archive retains the complete candidate tree, validity records, parent to child relationships, and within a run endpoints. Candidate freezing rules are given in Appendix~\ref{app:candidate-selection}. Original run directories and file checksums are retained in the accompanying reproduction manifest.

\begin{table}[t]
\caption{Candidate validity and final candidates of the search seed 42 source search.}
\centering
\scriptsize
\resizebox{\columnwidth}{!}{%
\begin{tabular}{lrrrl}
\toprule
Source & Stage I & Stage III & Unique seeds & Final candidates\\
\midrule
AutoSND-Q/S & 99.7\% & 98.0\% & 11 & Q: \texttt{cd9e3818033d}, S: \texttt{0ade8d3405c2}\\
\bottomrule
\end{tabular}}
\end{table}
This source search follows the unified configuration in Section~\ref{sec:search-settings} and Table~\ref{tab:config}. The four seed types and parent to child relationships follow the archive records.

\subsection{Proposal, Policy, Prompt, and Static Features}

\begin{table}[t]
\caption{Tracing relationships among Stage II proposals, policy, prompts, and code features.}
\centering
\scriptsize
\resizebox{\columnwidth}{!}{%
\begin{tabular}{llllrr}
\toprule
Candidate & Traced item & Proposal & Policy & Prompt & Stage I$\to$III\\
\midrule
Q & Adjacency cache/preconstruction & 75.0\% & Retained & 100\% & 0\%$\to$98.8\%\\
Q & Bounded two-hop & 100\% & Retained & 100\% & 0\%$\to$21.3\%\\
Q & Lazy heap & 100\% & Retained & 100\% & 100\%$\to$100\%\\
S & Bounded two-hop & 100\% & Retained & 100\% & 100\%$\to$100\%\\
S & Stage scheduling & 100\% & Retained & 100\% & 100\%$\to$99.0\%\\
\bottomrule
\end{tabular}}
\end{table}
The cache feature change of AutoSND-Q is accompanied by a decrease in median candidate runtime from 0.1319 s to 0.0667 s over five proxy graphs (\texttt{SynthPL500\_g25\_s11}, \texttt{SynthPL1000\_g22\_s17}, \texttt{SynthComm1000\_s23}, \texttt{SynthGridSW1200\_s31}, and \texttt{Powerlaw\_500}). The median static literal count of \texttt{.neighbors(} for AutoSND-S decreases from 6 to 4. The two candidate tracing chains jointly record implementation changes conditioned on the structural policy.

\subsection{Sources of Policy Fields}

\begin{table}[t]
\caption{Evidence source composition of compiled policy fields.}
\centering
\scriptsize
\begin{tabular}{lrrrrr}
\toprule
Candidate & Fixed & Fixed+proposal & Proposal only & Config./stats & Fields\\
\midrule
AutoSND-Q & 21.9\% & 26.6\% & 46.9\% & 4.7\% & 64\\
AutoSND-S & 15.8\% & 25.0\% & 55.3\% & 3.9\% & 76\\
\bottomrule
\end{tabular}
\end{table}
The source resolution and prompt coverage rates of all 140 policy fields are 100\%. The fixed validity envelope provides the safety boundary, while Stage I statistics and Stage II proposals provide data driven preferences. Including combined sources such as ``prior + proposal'' and ``statistics + proposal,'' fields involving proposals account for 75.0\% and 81.6\% of Q/S, respectively.

\subsection{System Level Whole Package Comparison}
\label{app:system-comparison}

\begin{table}[t]
\caption{System level comparison between complete AutoSND and \texttt{w/o Stage II structure guidance}.}
\centering
\scriptsize
\begin{tabular}{lrrl}
\toprule
Setting & Stage I & Stage III & Complete sequence coverage\\
\midrule
Complete AutoSND-Q/S & \textbf{99.7\%} & \textbf{98.0\%} & Q/S: \textbf{100\%}\\
w/o Stage II guidance & \textbf{99.7\%} & 95.0\% & Q/S: 66.7\% (8/12)\\
\bottomrule
\end{tabular}
\end{table}
This comparison tests the system level role of Stage II structural guidance in complete AutoSND. Separate analyses of the two key components are given in Section~\ref{sec:ablation}.

\subsection{Search Distribution and Lineage Endpoints}
\label{app:lineage-endpoints}

\begin{table}[t]
\caption{Candidate validity and final lineage improvements of AutoSND-Q/S.}
\centering
\scriptsize
\resizebox{\columnwidth}{!}{%
\begin{tabular}{lrrrll}
\toprule
Candidate & Stage I & Stage III & Depth & Seed$\to$final ANC@30\% (GCC) & Seed$\to$final runtime\\
\midrule
AutoSND-Q & 99.7\% & 98.0\% & 8 & 0.526735$\to$\textbf{0.516803} & 0.097842$\to$\textbf{0.074128}\\
AutoSND-S & 99.7\% & 98.0\% & 5 & 0.524955$\to$\textbf{0.516595} & 0.149705$\to$\textbf{0.091376}\\
\bottomrule
\end{tabular}}
\end{table}
Both final lineages in the same source search reduce ANC@30\% (GCC) and candidate runtime simultaneously. Together with the distribution expansion and budget concentration evidence in the main text, this result supports the interpretation of ``selectively deepening responsive lineages.''

\subsection{Evidence Scope}

This tracing provides evidence for three conclusions: structural proposals are written into subsequent prompts, cache recommendations correspond to changes in static cache features, and the Q/S tracing chains exhibit differentiated implementation distributions. Single component contributions are measured by the code ablations in Appendix~\ref{app:component-ablation}, the system level whole package comparison is given in Appendix~\ref{app:system-comparison}, search lineage endpoints are given in Appendix~\ref{app:lineage-endpoints}, and external large graph quality is validated by the strict complete sequence experiments in Appendix~\ref{app:large-reproduction}. Proposals, policy fields, prompt coverage, lineages, and final candidate determination records are jointly indexed by the Stage II tracing items in the accompanying reproduction manifest.

\begin{table}[t]
\caption{Complete sequence ANC (GCC) on three large real networks.}
\centering
\scriptsize
\begin{tabular}{lrrrr}
\toprule
Method & Facebook & YouTube & Flickr & Mean\\
\midrule
FINDER & 0.268434 & 0.023665 & 0.054554 & 0.115551\\
AutoSND-S & 0.265373 & 0.023095 & 0.055986 & 0.114818\\
AutoSND-Q & 0.264462 & 0.022129 & 0.055667 & 0.114086\\
\bottomrule
\end{tabular}
\end{table}

\begin{table}[t]
\caption{Candidate validity and candidate runtime during search. Candidate runtime is calculated only over unique valid candidates that pass syntax, interface, execution, and output checks. P95 is the 95th percentile of valid candidate runtime.}
\centering
\scriptsize
\resizebox{\columnwidth}{!}{%
\begin{tabular}{lrrrrl}
\toprule
Stage & Budget & Unique valid & Validity & Duplicates & Candidate runtime mean/med/P95/max\\
\midrule
Stage I & 300 & 241 & 80.33\% & 0 & 3.910/3.596/6.950/83.846\\
Stage III & 200 & 155 & 77.50\% & 1 & 0.302/0.196/0.622/5.873\\
Total & 500 & 396 & 79.20\% & 1 & --\\
\bottomrule
\end{tabular}}
\end{table}

\begin{table}[t]
\caption{Candidate validity of each Stage III branch.}
\centering
\scriptsize
\begin{tabular}{lrrrr}
\toprule
Branch & Budget & Unique valid & Validity & Duplicates\\
\midrule
Q & 60 & 52 & 86.67\% & 0\\
S & 60 & 55 & 91.67\% & 0\\
B & 50 & 28 & 56.00\% & 0\\
R & 30 & 20 & 66.67\% & 1\\
\bottomrule
\end{tabular}
\end{table}

\begin{table}[t]
\caption{ANC (GCC) and candidate runtime for each graph of frozen AutoSND-Q/S generated by DeepSeek-v4-flash. Lower ANC (GCC) is better. \texttt{TO} means that no valid complete sequence was returned within 600 s. The final row computes successful case means over the 11 graphs completed by both Q and S.}
\centering
\scriptsize
\resizebox{\columnwidth}{!}{%
\begin{tabular}{lrrrr}
\toprule
Network & Q ANC (GCC) & \shortstack{Q candidate\\runtime} & S ANC (GCC) & \shortstack{S candidate\\runtime}\\
\midrule
CEnew & 0.146992 & 0.182 & 0.125204 & 0.138\\
Collaboration & 0.129411 & 10.201 & 0.103090 & 7.864\\
condmat & 0.205561 & 365.695 & 0.139459 & 278.401\\
crime & 0.114899 & 0.382 & 0.130761 & 0.318\\
email & 0.244822 & 0.838 & 0.256840 & 0.633\\
Grid & 0.057460 & 12.166 & 0.078418 & 10.400\\
hamster & 0.346180 & 3.353 & 0.191045 & 2.373\\
HepPh & 0.400846 & 107.258 & 0.297339 & 70.064\\
Yeast & 0.188944 & 3.422 & 0.185918 & 2.581\\
HI-II-14 & 0.081579 & 7.966 & 0.073919 & 7.358\\
Digg & 0.108742 & 528.220 & 0.100752 & 480.684\\
Gnutella31 & TO & 600.000 & TO & 600.000\\
\textbf{Mean over 11 successful graphs} & \textbf{0.184130} & \textbf{94.517} & \textbf{0.152977} & \textbf{78.256}\\
\bottomrule
\end{tabular}}
\end{table}

\section{Complete Experimental Results}
\label{app:fullresults}

\subsection{Complete Experimental Results with Six Decimal Places}
\label{app:full-six-decimal}

This section retains unscaled six decimal results following the statistical protocol in Section~\ref{sec:evaluation-metrics}. Tables~\ref{tab:fullahd}--\ref{tab:fullnative} summarize 17 medium sized graphs, followed by the three large real networks. Membership and data roles of the 12 main graphs are given in Appendix~\ref{app:data}. \texttt{TO} and ``--'' have the same meanings as in Table~\ref{tab:maincomparison}. Method specific candidate runtime sources and denominators are given in Appendix~\ref{app:data}.

\section{Supplementary Experiments with DeepSeek-v4-flash as the Generation Model}

\subsection{Model and Experimental Settings}

This experiment replaces only the generation model with DeepSeek-v4-flash. The remaining search procedure, candidate budget, proxy graph, branch configuration, and evaluation protocol follow Section~\ref{sec:setup} and Appendix~\ref{app:search}.

\subsection{Candidate Validity and Candidate Runtime}

\begin{table*}[t]
\caption{Complete sequence results of AHD methods, Codex, and AutoSND. Columns CEn, Col, con, cri, ema, Gri, GrQ, ham, Hep, PH, PL5, Yea, HI, Dig, Enr, Gnu, and Epi denote the 17 graphs in the order listed in Appendix~\ref{app:dataset-sizes}.}
\label{tab:fullahd}
\centering\tiny
\resizebox{\textwidth}{!}{%
\begin{tabular}{lrrrrrrrrrrrrrrrrrrrr}
\toprule
Method&CEn&Col&con&cri&ema&Gri&GrQ&ham&Hep&PH&PL5&Yea&HI&Dig&Enr&Gnu&Epi&Mean&Valid&Candidate runtime/Cov.\\
\midrule
FunSearch&.267128&.091413&.110662&.110255&.211218&.034877&.089890&.167517&.165251&.168652&.092248&.125859&.112900&.087007&TO&.116138&TO&.133352&98.4\%&79.232599/12\\
Clade-AHD&.096760&.075438&.114755&.110360&.211977&.038874&.075438&.152755&.144873&.152755&.095648&.137255&.055857&.085990&TO&.117751&TO&.111887&72.2\%&93.905165/12\\
ERA&.094820&.085223&.108104&.110472&.215713&.040204&.084843&.148643&.151547&.148499&.095884&.120746&.056465&.085772&TO&.116489&TO&.111183&79.6\%&29.293214/12\\
MCTS-AHD&.096940&.101827&.124535&.111563&.220889&.044419&.101826&.174894&.178686&.174894&.096144&.135917&.056927&.086185&TO&.117669&TO&.120871&75.6\%&94.278312/12\\
AlphaEvolve&.098753&.107088&.130063&.111160&.223071&.046801&.107242&.189217&.185847&.189217&.097552&.142982&.057862&.086578&.046265&.119557&.051446&.124915&89.4\%&9.012329/12\\
LLM4CN&.125019&.087214&TO&.124332&.239847&.059348&.087214&.168357&.147255&.168357&.106148&.132484&.063416&TO&TO&TO&TO&--&93.8\%&76.368434/9\\
EoH&.100083&.107316&.130014&.111440&.222525&.045807&.107316&.188005&.186957&.188005&.096284&.141913&.057542&.086066&TO&.117981&TO&.124637&95.4\%&58.059054/12\\
ReEvo&.099192&.100212&.124857&.112867&.223775&.048486&.100212&.181801&TO&.181801&.098144&.136259&.059961&.087665&TO&.122905&TO&--&60.9\%&23.307139/11\\
HiFo-Prompt&.100161&.115262&.137305&.111822&.226405&.049838&.115262&.199783&TO&.199783&.096384&.148732&.059975&.087479&TO&.123266&TO&--&100.0\%&16.538775/11\\
HSEvo&.094211&.103439&.117536&.111555&.218816&.043205&.103439&.169202&TO&.169202&.095332&.159107&.057138&TO&TO&TO&TO&--&67.7\%&47.421578/9\\
Codex-Evo&.096779&.100513&TO&.107060&.218678&.044007&.100499&.173917&TO&.173917&.093044&.133363&.055982&TO&TO&TO&TO&--&23.6\%&15.246338/8\\
Codex-Direct&.089333&TO&TO&.108176&.215086&TO&TO&TO&TO&TO&.093732&TO&TO&TO&TO&TO&TO&--&60.0\%&71.675619/3\\
AutoSND-S&.097973&.074274&.108292&.112937&.216998&.045836&.075544&.149873&.136693&.148882&.099736&.114185&.055653&.087343&.040742&.111402&.049859&.109288&99.0\%&2.062209/12\\
AutoSND-Q&.096589&.077088&.107660&.109896&.216723&.038702&.078610&.151921&.141108&.149001&.092768&.120995&.055721&.085745&.040166&.116563&.048906&.109893&99.0\%&2.861835/12\\
\bottomrule
\end{tabular}}
\end{table*}

\begin{table*}[t]
\caption{Complete sequence results of native and deterministic methods, FINDER, and AutoSND. Graph abbreviations follow Table~\ref{tab:fullahd}. ANC (GCC) and candidate runtime for HDA, GND, MinSum, and BPD on the 12 main graphs come from rule preserving accelerated reevaluation. FINDER candidate runtime comes from same source materials and is averaged over 11 main test graphs with candidate runtime records.}
\label{tab:fullnative}
\centering\tiny
\resizebox{\textwidth}{!}{%
\begin{tabular}{lrrrrrrrrrrrrrrrrrrr}
\toprule
Method&CEn&Col&con&cri&ema&Gri&GrQ&ham&Hep&PH&PL5&Yea&HI&Dig&Enr&Gnu&Epi&Mean&Candidate runtime/Cov.\\
\midrule
HDA&.108923&.107122&.130348&.113267&.225433&.055091&.107293&.186873&.184679&.186873&.101944&.138864&.057751&.088145&.045200&.114626&.052000&.125927&.348331/12\\
CoreHD&.100522&.107598&.130560&.116537&.225810&.050584&.107664&.189973&.184745&.189973&.096744&.140055&.057069&.086178&.051400&.112867&.054600&.125208&.342789/12\\
DC&.118245&.131569&.147339&.123925&.250634&.065187&.132384&.200132&.230020&.200132&.112664&.195806&.069441&.096719&.046382&.129402&.057707&.146535&.110224/12\\
CI&.128722&.136768&.193881&.152194&.270623&.079483&.136774&.210003&.268728&.210003&.149236&.236880&.089300&.108532&.044500&.144689&.050600&.168317&.792828/12\\
KCore&.185162&.228222&.273978&.260175&.294754&.263077&.232995&.257298&.283194&.257298&.216288&.289314&.108381&.109660&.078216&.143463&.066485&.224723&.309684/12\\
CLUC&.494238&.460136&.486775&.462590&.427748&.294482&.460335&.483401&.478446&.483401&.316760&.397639&.154982&.131580&.092204&.155616&.077079&.368969&.788001/12\\
GND&.090366&.106951&.126458&.119639&.223093&.041155&.104065&.142459&.184211&.147513&.102464&.120872&.060647&.087153&.045600&.120463&.053700&.118622&14.550765/12\\
MinSum&.092462&.095759&.123349&.109358&.221381&.043611&.095849&.174259&.172357&.174259&.094512&.136502&.055545&.085497&.047700&.113792&.060400&.118656&14.043380/12\\
BPD&.092842&.107686&.127608&.110695&.223648&.044203&.107688&.180611&.177003&.180611&.094792&.143057&.059232&.087514&.061800&.121571&.074800&.122973&11.763963/12\\
FINDER&.107027&.114333&.130191&.109873&.222728&.050041&.114350&.189229&.198225&.189231&.096300&.134631&.055400&.086600&.043000&.111000&.049900&.125773&6.947583/12\\
AutoSND-S&.097973&.074274&.108292&.112937&.216998&.045836&.075544&.149873&.136693&.148882&.099736&.114185&.055653&.087343&.040742&.111402&.049859&.109288&2.062209/12\\
AutoSND-Q&.096589&.077088&.107660&.109896&.216723&.038702&.078610&.151921&.141108&.149001&.092768&.120995&.055721&.085745&.040166&.116563&.048906&.109893&2.861835/12\\
\bottomrule
\end{tabular}}
\end{table*}

The 500 candidate calls produce 396 unique valid candidates, for an overall validity rate of 79.20\%. The mean candidate runtime of valid Stage III candidates decreases from 3.910 s in Stage I to 0.302 s, and P95 decreases from 6.950 s to 0.622 s. Candidate validity in the Q and S branches reaches 86.67\% and 91.67\%, respectively. These results show that under the DeepSeek-v4-flash generation setting, the candidate distribution after Stage II shifts markedly toward lighter local update implementations.

\subsection{Complete Sequence ANC (GCC), Candidate Runtime, and Result Analysis}

On the 11 commonly completed networks, S obtains lower ANC (GCC) on eight graphs and shorter candidate runtime on all 11 graphs. Its mean ANC (GCC) is 16.9\% lower than Q, and its mean candidate runtime is 17.2\% lower. Q obtains lower ANC (GCC) on crime, email, and Grid. The two candidates therefore show complementary external performance, with S providing the better overall quality and candidate runtime operating point. Together with the candidate validity and candidate runtime distributions, this experiment shows that the three stage AutoSND process can continue to form executable final candidates under the DeepSeek-v4-flash generation setting.

\end{document}